\documentclass[journal]{IEEEtran}
\ifCLASSINFOpdf \else \fi
\pdfoutput=1

\usepackage[cmex10]{amsmath}
\usepackage{amsmath}
\usepackage{graphics}
\usepackage{graphicx}
\usepackage{caption}
\usepackage{color}
\usepackage{soul}
\usepackage{booktabs}
\usepackage{adjustbox}
\usepackage{tabularx}
\usepackage{multirow}
\usepackage{array}
\usepackage{amsfonts}
\usepackage{algorithm}
\usepackage{algorithmic}
\setstcolor{red}
\usepackage{tabularray}
\usepackage{mathrsfs}

\usepackage{array}
\ifCLASSOPTIONcompsoc
 \usepackage[caption=false,font=normalsize,labelfont=sf,textfont=sf]{subfig}
\else
 \usepackage[caption=false,font=footnotesize]{subfig}
\fi
\usepackage{url}

\usepackage[noadjust]{cite}

\begin{document}

\title{Meta-learning in healthcare: A survey}

\author{{Alireza Rafiei, Ronald Moore, Sina Jahromi, Farshid Hajati, Rishikesan Kamaleswaran}% <-this % stops a space
\thanks{A Rafiei and R Moore are with the Department of Computer Science at Emory University, Atlanta, GA, USA. S Jahromi is with the Department of Mechatronics Engineering, University of Tehran, Tehran, Iran. F Hajati is with the School of Science and Technology, University of New England, NSW, Australia. R Kamaleswaran is with Emory University School of Medicine, Atlanta, GA, USA. R Moore and S Jahromi had equal contributions.}
}

\markboth{Meta-learning in healthcare: A survey}%
{Shell \MakeLowercase{\textit{et al.}}: Bare Demo of IEEEtran.cls for Journals}

\maketitle

% As a general rule, do not put math, special symbols or s
\begin{abstract}
As a subset of machine learning, meta-learning, or learning to learn, aims at improving the model's capabilities by employing prior knowledge and experience. A meta-learning paradigm can appropriately tackle the conventional challenges of traditional learning approaches, such as insufficient number of samples, domain shifts, and generalization. These unique characteristics position meta-learning as a suitable choice for developing influential solutions in various healthcare contexts, where the available data is often insufficient, and the data collection methodologies are different. This survey discusses meta-learning broad applications in the healthcare domain to provide insight into how and where it can address critical healthcare challenges. We first describe the theoretical foundations and pivotal methods of meta-learning. We then divide the employed meta-learning approaches in the healthcare domain into two main categories of multi/single-task learning and many/few-shot learning and survey the studies. Finally, we highlight the current challenges in meta-learning research, discuss the potential solutions, and provide future perspectives on meta-learning in healthcare. \\
\par 
\textit{\textbf{Index terms}}-Meta-leaning, healthcare, few-shot learning, multi-task learning, clinical decision support system, automated medicine
\end{abstract}

\IEEEpeerreviewmaketitle

\section{Introduction}
\IEEEPARstart{F}{ueled} by the surge in the collection of diverse data, coupled with advancements in computational models and algorithms, artificial intelligence (AI) techniques have been striving to establish a strong foothold in healthcare over the past decade  \cite{he2019practical,malik2019overview,hamet2017artificial}. This burgeoning trend has fostered a growing interest in the deployment of innovative data analysis methods and machine learning (ML) techniques across a range of healthcare applications \cite{johnson2016machine,wainberg2018deep,dash2019big,nayyar2021machine}. As a specialized area within ML, meta-learning, or learning-to-learn, has recently gained significant attention due to its impressive theoretical and practical advancements, making it a primary choice for numerous applications \cite{tian2022meta,huisman2021survey,hospedales2021meta}.

\par In the meta-learning paradigm, the objective transcends beyond learning a specific task. Instead, algorithms are designed to understand the process of learning itself, equipping them with the capability to swiftly adapt to new tasks or changes, often based on limited data or even a single example \cite{tian2022meta,huisman2021survey}. The potent capacity for rapid adaptation by utilizing prior knowledge and experience differentiates meta-learning from more traditional ML methodologies. In essence, meta-learning enables learners to adapt adeptly to changing conditions by identifying clues that inform a learning process. This strategy potentially leads to a more resource and time-efficient learning process, as it minimizes the number of experiments needed to discern patterns \cite{vanschoren2018meta}. Furthermore, meta-learning can find optimal solutions both without necessitating the availability of a mathematical model and with integrating modeling of a problem \cite{finn2017model,santoro_meta-learning_2016}. 

\par These unique characteristics position meta-learning as an enticing solution for building efficient models in various healthcare-related scenarios. The healthcare domain frequently entails complex decision-making processes often characterized by extended time frames, diverse and multimodal data, and patient-specific outcomes. A learning model in this space needs to be able to learn and adapt quickly, and meta-learning provides this capability. The design of meta-learning algorithms to generalize from past experience leads to effectively adapting to new tasks and concepts \cite{vanschoren2019meta,wang2019survey}. Thus, meta-learning models can handle domain shifts and transfer knowledge from one medical condition to another. This ability not only avails the opportunity to utilize data and experience from different medical conditions to develop a learning model for the targeted condition, but it also offers more personalized care by considering individual levels.

\par Despite the initial success of common deep learning (DL) models in the healthcare domain, they typically perform well on a single task \cite{zhou2021review,si2021deep}. Meta-learning models, however, prove beneficial both in multi-task scenarios, where task-agnostic knowledge is garnered from a suite of tasks to enhance the learning of new tasks within that suite, and in single-task scenarios, where a single problem is continually solved and refined solutions for a single problem over numerous episodes \cite{hospedales2021meta,zou2022meta}. This multi-task learning capability can enable a more comprehensive understanding of the complex interrelations and dependencies between various healthcare conditions and provide a more robust model. Moreover, the efficacy of meta-learning shines in situations where data availability is scarce, a frequent occurrence in the healthcare domain. Where only limited data is available for training, the meta-learning models demonstrate their prowess. Unlike ML and DL models that require large volumes of data to produce accurate predictions, meta-learning models can learn from a small number of examples and still deliver reliable results. These abilities render meta-learning approaches more attractive than many existing methods in the healthcare domain.

\begin{table}[h]
\centering
\caption{Summary of acronyms in meta-learning.}
\begin{tabular}{l|l} 
\hline \hline
\textbf{Acronym}  & \textbf{Description}                                                 \\ 
\hline
AI       & Artificial
  Intelligence                                   \\ 
\hline
AMGNN    & Auto-Metric
  Graph Neural Network                          \\ 
\hline
ANIL     & Almost No Inner
  Loop                                      \\ 
\hline
BFA      & Bayesian Fast
  Adaptation                                  \\ 
\hline
CAD      & Computer-Aided
  Diagnosis                                  \\ 
\hline
CE       & Cross-Entropy                                               \\ 
\hline
CL       & Curriculum
  Learning                                       \\ 
\hline
CNF      & Continuous
  Normalizing Flow                               \\ 
\hline
CNN      & Convolutional
  Neural Network                              \\ 
\hline
CNP      & Conditional Neural Processes                             \\ 
\hline
ConvLSTM & Convolutional
  Long Short-Term Memory                      \\ 
\hline
CRNN     & Convolutional
  Recurrent Neural Network                    \\ 
\hline
CRPM     & Clinical Risk
  Prediction Model                            \\ 
\hline
DASNet   & Dual Adaptive
  Sequential Network                          \\ 
\hline
DDI      & Drug-Drug
  Interaction                                     \\ 
\hline
DL       & Deep Learning                                               \\ 
\hline
DSC       & Dice Similarity Coefficient                                             \\ 
\hline
ES       & Evolution
  Strategies                                      \\ 
\hline
FOMAML   & First-Order
  Model-Agnostic Meta-Learning                  \\ 
\hline
FTL      & Follow The
  Leader                                         \\ 
\hline
FTML     & Follow The
  Meta-Leader                                    \\ 
\hline
GCN     & Graph Convolutional Networks                                    \\ 
\hline
GET      & Graph-Enhanced
  Transformer                                \\ 
\hline
GP       & Genetic
  Programming                                       \\ 
\hline
GRNN     & Generalized
  Regression Neural Network                     \\ 
\hline
GTE      & Graph Enhanced
  Transformer~                               \\ 
\hline
IFLF     & Invariant
  Feature Learning Framework                      \\ 
\hline
iMAML    & Implicit
  Mode-Agnostic Meta-Learning                      \\ 
\hline
LRUA     & Least Recently
  Used Access                                \\ 
\hline
ISGNN    & Interpretable
  Structure-Constrained Graph Neural Network  \\ 
\hline
LSTM     & Long Short-Term
  Memory                                    \\ 
\hline
MAML     & Mode-Agnostic
  Meta-Learning                               \\ 
\hline
MANN     & Memory-Augmented
  Neural Network                           \\ 
\hline
MetaNet  & Meta Network                                                \\ 
\hline
ML       & Machine Learning                                            \\ 
\hline
MLP      & Multi-Layer
  Perceptron                                    \\ 
\hline
MMAML    & Multimodal
  Mode-Agnostic Meta-Learning                    \\ 
\hline
MMD      & Maximum Mean
  Discrepancy                                  \\ 
\hline
MO       & Molecular
  Optimization                                    \\ 
\hline
MORF     & Meta-Ordinal
  Regression Forest                            \\ 
\hline
MSE      & Mean Squared
  Error                                        \\ 
\hline
NGL      & Neural Graph
  Learning                                     \\ 
\hline
NLP      & Natural Language
  Processing                               \\ 
\hline
NTM      & Neural Turing
  Machine                                     \\ 
\hline
OCTRC    & Oregon Clinical
  and Translational Research Center         \\ 
\hline
ProtoNet & Prototypical
  Network                                      \\ 
\hline
PSO      & Particle Swarm
  Optimization                               \\ 
\hline
PTB      & Prioritization
  Task Buffer                                \\ 
\hline
QSAR     & Quantitative
  Structure-Activity Relationships             \\ 
\hline
r2d2     & Ridge Regression
  Differentiable Discriminator             \\ 
\hline
RL       & Reinforcement
  Learning                                    \\ 
\hline
RN       & Relation Network                                            \\ 
\hline
RMAML    & Robust
  Mode-Agnostic Meta-Learning                        \\ 
\hline
RR       & Ridge Regression                                            \\ 
\hline
SignSGD  & Sign-based
  Stochastic Gradient Descent                    \\ 
\hline
SMeta    & Side-aware
  Meta-learning                                  \\ 
\hline
SNN      & Siamese neural network                                  \\
\hline
TAML     & Task-Agnostic
  Meta-Learning                               \\
\hline \hline
\end{tabular}
\end{table}

\par To date, numerous theoretical and experimental studies have successfully incorporated meta-learning approaches into various healthcare contexts. However, despite these advancements, the application of meta-learning in healthcare remains in its nascent stages and has not yet garnered significant attention within the broader research community. As the first comprehensive survey addressing meta-learning within this domain, our paper aims to bridge a gap in the existing literature by synthesizing recent advancements and their practical applications. We provide new insights into how meta-learning techniques can be tailored to overcome unique challenges associated with medical data, such as heterogeneity, incompleteness, and sparsity. Additionally, our survey serves as an accessible entry point for individuals new to this field, offering a clear vision of the path ahead by detailing the fundamentals of meta-learning techniques and illustrating their relevance to various healthcare scenarios. This enables researchers and practitioners to systematically select appropriate methodologies for specific tasks. The paper also explores open areas, essential future research avenues, and the potential impacts of meta-learning in healthcare, delivering valuable conclusions that guide both ongoing research and practical implementations—areas not extensively covered in existing surveys. The rest of the paper is organized as follows: Section II gives a big picture of the utilities of meta-learning in healthcare and introduces the discussed categories in this paper. Section III provides a summary of the theory and pivotal techniques in meta-learning research. Section IV presents multi- and single-task learning and Section V discusses many- and few-shot learning meta-learning-based methods in healthcare. Additionally, Section VI discusses the development highlights of a meta-learning model. Section VII provides a comprehensive discussion of bias, model validation, generalizability, interoperability, and other challenges and open issues in this domain. Eventually, section VIII describes potential future directions for applying meta-learning in healthcare. For convenience, the main acronyms in meta-learning and healthcare used in this paper were summarized in Table I and Table II, respectively.

\begin{table}[t]
\centering
\caption{Summary of acronyms in healthcare.}
\begin{tabular}{p{1.3cm}|p{6.5cm}} 
\hline \hline
\textbf{Acronym} & \textbf{Description}                 \\ 
\hline
AF               & Atrial Fibrillation                  \\ 
\hline
ARDS             & Acute Respiratory Distress Syndrome  \\ 
\hline
ASD              & Autism Spectrum Disorder             \\ 
\hline
BG               & Blood Glucose                        \\ 
\hline
Bio-Z          & Bio-impedance~                       \\ 
\hline
CT               & Computed Tomography                  \\ 
\hline
CVD              & CardioVascular Disease               \\ 
\hline
CXR              & Chest X-Ray                          \\ 
\hline
DDI              & Drug-Drug Interaction                \\ 
\hline
ECG              & Electrocardiogram                    \\ 
\hline
EEG              & Electroencephalogram                 \\ 
\hline
EHR              & Electronic Health Record             \\ 
\hline
HAR              & Human Activity Recognition           \\ 
\hline
ICU              & Intensive Care Unit                  \\ 
\hline
IDF              & International Diabetes Federation    \\ 
\hline
LOS              & Length Of Dtay                       \\ 
\hline
MCI              & Mild Cognitive Impairment            \\ 
\hline
NGS              & Next-Generation Sequencing           \\ 
\hline
PM               & Peritoneal Metastases                \\ 
\hline
PsP              & Pseudo-Progression                   \\ 
\hline
RRIs             & Residue-Residue Interactions         \\ 
\hline
SCD              & Subjective Cognitive Decline         \\ 
\hline
SER              & Speech Emotion Recognition           \\ 
\hline
SMC              & Subjective Memory Complaint          \\ 
\hline
TBI              & Traumatic Brain Injury               \\ 
\hline
TSS              & Transcription Start Sites            \\ 
\hline
TTP              & True Tumor Progression               \\ 
\hline
VA               & Ventricular Arrhythmias              \\ 
\hline
VF               & Ventricular Fibrillation             \\ 
\hline
WES              & Whole-Exome Sequencing               \\ 
\hline
HIV              & Human Immunodeficiency Virus         \\ 
\hline
WGD              & Whole-Genome Doubling                \\ 
\hline
WGS              & Whole-Genome Sequencing              \\
\hline \hline
\end{tabular}
\end{table}

\section{Utilities of meta-learning in healthcare}
Owing to the progress in ubiquitous monitoring, advanced censoring, and measurement techniques, healthcare-related data can be generated in a wide range of formats, such as time series measurements, physiological wave-forms, images, protein structures, and free-text notes, providing opportunities for developing ML techniques in critical care \cite{provost2022health}. Nevertheless, healthcare data are typically incomplete, deficient, noisy, and biased \cite{johnson2016machine, dash2019big}. Hence, analyzing these data to discern the appropriate pattern is the premier challenge of automated systems. The systematic development of learning to learn based algorithms in the ML field is relatively new and more advanced models are being developed. The unique abilities of these algorithms, such as the ability to deliver high performance while a limited number of data is available, handle domain shifts, generalizability, and use of prior knowledge and experience, make them different from traditional ML approaches. Given the characteristics of meta-learning methods and healthcare data, it becomes an appealing solution for tackling healthcare challenges. 

\par
\begin{figure*}[h]
  \centering
  \includegraphics[width=1\textwidth]{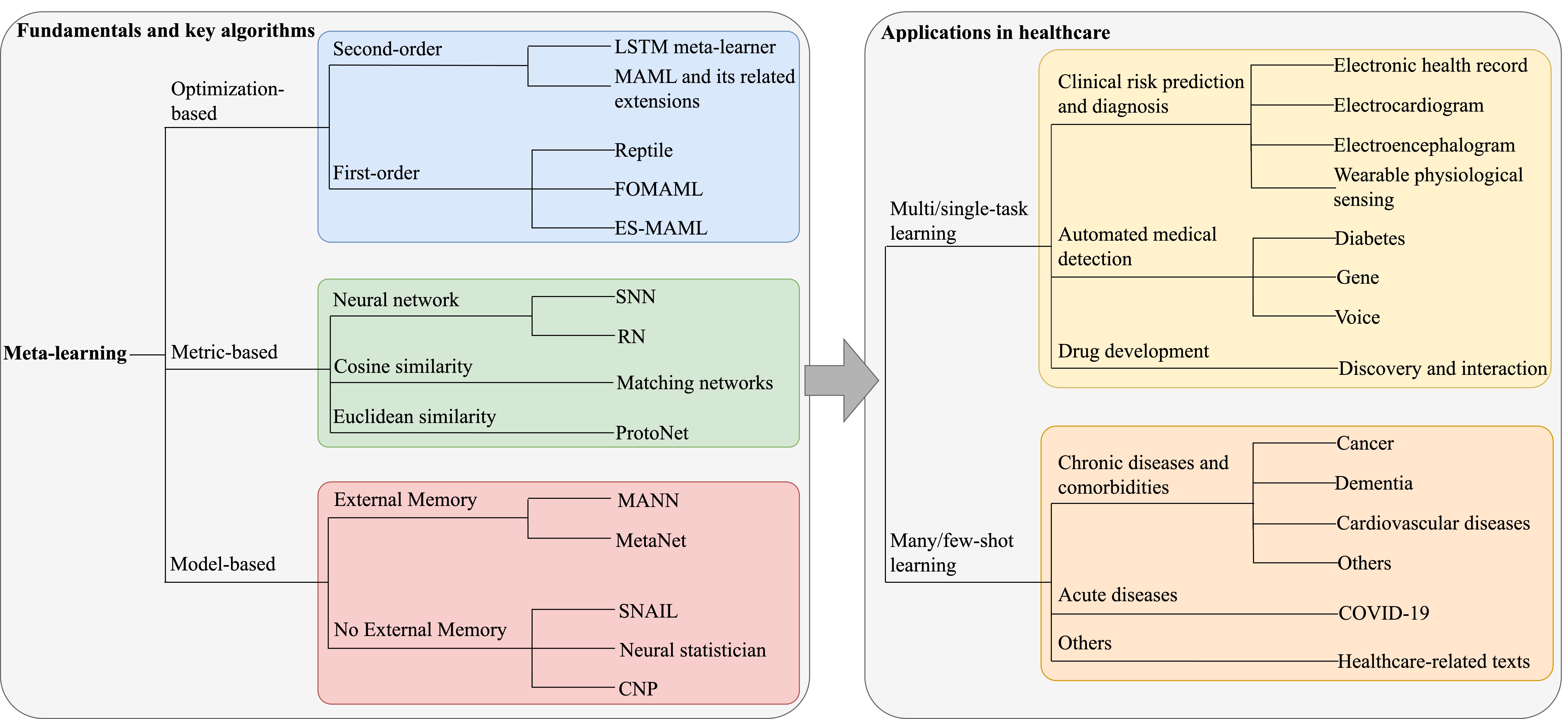}
  \caption{The conceptual diagram of the fundamentals and key algorithms (left), and application domains of meta-learning in healthcare (right).}
  \label{fig:diag}
\end{figure*}

 This paper explores meta-learning strategies to elucidate their utility and influence across a wide range of healthcare conditions and data resources. Since the terms meta-learning, meta-learner, and meta-model have been used for various methodological schemes, the primary criterion for including publications in this survey is that the research should present the concept of \textit{learning to learn} as meta-learning. This is done by first reviewing the abstract and, if necessary, screening the entire paper. We defined the scope by identifying key areas where meta-learning could significantly impact healthcare. We categorize the applications of meta-learning approaches into two main sections based on the available research studies: multi/single-task learning and many/few-shot learning. The multi/single-task learning section consists of clinical risk prediction and diagnosis, automated medical detection, and drug development subsections. The many/few-shot learning section encompasses image-based methods, including chronic disease and comorbidities and acute diseases, as well as healthcare-related text approaches. Figure \ref{fig:diag} presents a conceptual diagram that provides an overview of discussed theoretical topics in this survey and the application of meta-learning in healthcare based on the mentioned two broad domains. The further categorization of these two sections is rooted in fundamental distinctions that are medically relevant and related to the input modalities employed. The included publications in this survey have been screened and selected using different combinations of the mentioned keywords in Figure \ref{fig:diag}, as well as other healthcare disorders and diseases, from PubMed, IEEE Xplore, and Google Scholar academic databases without any restrictions on data type, performance, meta-learning methodologies, and publisher. The studies discussed in the following sections have utilized algorithms from one of the three principal categories of meta-learning methods, as outlined in Section III. Additionally, ones with considerable contributions to the applications of meta-learning in healthcare in multi/single-task learning and many/few-shot learning categories are summarized in Table III and Table IV, respectively.

\section{Fundamentals and key algorithms in meta-learning}
The general definition of meta-learning is difficult, but they fundamentally can be considered as task distribution and bi-level optimization approaches \cite{hospedales2021meta,zou2022meta}. From the task distribution point of view, the primary goal of meta-learning is to enable a model to extend its performance to a set of tasks. We define the generalizability of the model as its ability to rapidly achieve exceptional performance on an unseen task from only a few examples. Given a task distribution $p(T)$, where $T = \{D, L\}$ represents a given task with dataset $D$ and loss function $L$, and model parameters $\omega$, the task distribution problem can subsequently be defined as:

\begin{equation}
    \min _\omega \underset{T \sim p(T)}{\mathbb{E}} L(D, \omega)
\end{equation}

In this sense, meta-learning aims to minimize the expected loss across a set of tasks, thereby increasing the generalizability of the model to unseen tasks. Meta-learning can also be viewed as a bi-level optimization problem, where one optimization problem contains another optimization problem as one of its constraints. The inner optimization problem is modeled using standard DL objective functions such as cross-entropy (CE) or mean squared error (MSE). However, the outer optimization problem is designed to contain a meta-learning objective. Given a set of \textit{M} source tasks $\mathscr{D}_{\text {source }}=\left\{\left(D_{source }^{train}, D_{source}^{val}\right)^{(i)}\right\}_{i=1}^M$, an inner objective $L_{task}$, and an outer objective $L_{meta}$, the bi-level optimization problem can thus then be defined as:

\begin{equation}
    \begin{aligned}
        \omega^* & =\underset{\omega}{\mathrm{argmin}} \sum_{i=1}^M L_{meta}\left(\theta^{*(i)}(\omega), \omega, D_{source}^{val}(i)\right) \\
        \text { s.t. } \theta^{*(i)}(\omega) & =\underset{\theta}{\mathrm{argmin}} \: L_{task}\left(\theta, \omega, D_{source}^{train(i)}\right)
    \end{aligned}
\end{equation}

where $\omega$ represents the model weights updated by the outer loop and $\theta^{*(i)}(\omega)$ are the weights for task $i$ updated in the inner loop. From the equation, it can be seen that the training set for task $i$ $D_{source}^{train(i)}$ is used to train the weights in the inner objective, while the validation set for task \textit{i} $D_{source}^{val(i)}$ is used to update the weights in the outer objective.

\par Following the taxonomy proposed by \cite{zou2022meta}, meta-learning methods fall into 3 broad categories: optimization-based, metric-based, and model-based. We further elucidate these three categories and their key algorithms in the following section.

\subsection{Optimization-based Models}
In optimization-based meta-learning, the goal is to design an optimization problem that converges quickly during training using a small number of samples for each task while also maintaining steady generalizability.

\subsubsection{LSTM Meta-learner}
The LSTM meta-learner uses an LSTM as a base learner to learn the appropriate update rule to train neural network classifier \cite{Ravi2016OptimizationAA}. During meta-training, the classifier uses the shared parameters from the LSTM meta-learner to make predictions for a given batch of training data. The classifier then returns the subsequent loss and gradients to the LSTM meta-learner so that it can suggest the appropriate parameters for the classifier. During testing, the classifier and its parameters are evaluated on the test data to generate a loss that is used to train the meta-learner. 

\subsubsection{Model-agnostic Meta-learning}
\par The most notable meta-representation algorithm is the model-agnostic meta-learning (MAML) algorithm \cite{finn2017model}. This algorithm learns a generalized representation of a distribution of tasks by finding model parameters that are sensitive to changes in a specific task with respect to the task's loss gradient. Given a model $f$, model parameters $\theta$, and a batch of tasks $T \sim p(T)$, for each task, $K$-shot learning is performed where $K$ samples from the task dataset are selected and used for training and feedback is gathered from the task loss function $L_{T_{i}}$. This feedback is used to update the model parameters relative to each task and is given by the equation

\begin{equation}
    \theta '_{i} = \theta - \gamma \nabla_{\theta} L_{T_{i}}(g_{\theta})
\end{equation}

where $\gamma$ is a hyperparameter for the step size used for the task gradients. After computing $\theta ' _{i}$ for each task, the model's parameters are updated using each task's adapted parameters and their respective loss function. This is represented by

\begin{equation}
    \theta \leftarrow \theta - \delta \nabla_{\theta} \sum_{T_{i} \sim p(T)} L_{T_{i}}(g_{\theta '_{i}})
\end{equation}

where $\delta$ is a hyperparameter for the step size used for the meta gradient. The MAML algorithm is outlined in detail in Algorithm \ref{alg:maml}.

\begin{algorithm}
    \caption{MAML}
    \label{alg:maml}
    \begin{algorithmic}
        \REQUIRE $p(T)$: distribution over tasks
        \REQUIRE $\gamma$, $\delta$: step size hyperparameters
        \STATE randomly initialize $\theta$
        \WHILE{not done}
        \STATE Sample batch of tasks $T_{i} \sim p(T)$
        \FORALL{$T_{i}$}
            \STATE Evaluate $\nabla_{\theta} L_{T_{i}}(g_{\theta})$ with respect to \textit{K} examples
            \STATE $\theta '_{i} = \theta - \gamma \nabla_{\theta} L_{T_{i}}(g_{\theta})$
        \ENDFOR
        \ENDWHILE
        \STATE $\theta \leftarrow \theta - \delta \nabla_{\theta} \sum_{T_{i} \sim p(T)} L_{T_{i}}(g_{\theta '_{i}})$
    \end{algorithmic}
\end{algorithm}

\subsubsection{First-order Model-Agnostic Meta-learning Extensions}
\par Many variants of MAML have been developed to further its success in creating generalizable representations. The most notable issue with the original MAML algorithm is the heavy computational costs associated with computing the gradients for the second-order derivatives. As a result, subsequent research has focused on developing MAML alternatives using first-order derivatives. Reptile is an approximation of the first-order MAML (FOMAML) algorithm \cite{nichol_first-order_2018, nichol2018reptile}. FOMAML and Reptile both use stochastic optimization within their inner loops. However, FOMAML uses the inner loop optimization to update the model parameters, while Reptile treats this optimization as a gradient and plugs it back into the gradient descent optimizer used for the inner loop. Evolution strategies MAML (ES-MAML) was another first-order MAML algorithm developed to address the estimating second derivatives problems on stochastic policies using backpropagation \cite{song2019maml}.

\subsubsection{Second-order Model-Agnostic Meta-learning Extensions}
\par There has also been research conducted to reduce second-order derivative costs without resorting to first-order derivatives. Raghu et al. \cite{raghu_rapid_2020} improved upon the MAML model with the almost no inner loop (ANIL) model. The authors argued that the MAML model does not gain significant knowledge from the inner loop updates. As a result, they remove the inner loop updates and only make parameter changes to the task-specific head. The ANIL model achieved comparable performance to the MAML model with significantly less training time. MAML++ is a faster and more memory-efficient version of MAML that incorporates multi-step loss optimization, derivative-order annealing, per-step batch normalization running statistics, per-step batch normalization weights and biases, learning per-layer per-step learning rates and gradient directions, and cosine annealing of the meta-optimizer learning rate \cite{antoniou_how_2019}.

\par Other MAML-related research has concentrated on stabilizing the gradients during training. Implicit MAML (iMAML) which uses implicit differentiation for the inner loop optimization of the MAML algorithm \cite{rajeswaran_meta-learning_2019}. By computing the inner loop meta-gradient through implicit differentiation, the outer loop meta-gradient computation becomes independent from the choice of inner loop optimizer, which results in a faster and more memory-efficient version of MAML. The ridge regression differentiable discriminator (r2d2) model uses a differentiable ridge regression (RR) layer in the inner loop to learn task-specific features \cite{bertinetto_meta-learning_2019}. In the outer loop, the network weights and RR layer hyperparameters are shared across the episodes, which improves the learning aptitude of the RR learner. The task-agnostic meta-learning (TAML) model updates the meta-learning parameters in the outer loop by adding a regularization term to the standard MAML meta-learning objective function that minimizes the entropy for each task \cite{jamal_task_2019}. Sign-MAML uses sign-based stochastic gradient descent (SignSGD) with the MAML algorithm \cite{fan_sign-maml_2021}. The gradient in the inner-loop optimization problem of MAML is unrolled and then solved with SignSGD.

\par Other researchers have developed MAML models using probabilistic methods. Finn et. al (2018) proposed a probabilistic latent model for incorporating priors and uncertainty in few-shot learning (PLATIPUS) \cite{finn_probabilistic_2018}. This probabilistic version of MAML utilizes a graphical representation of the model to account for the uncertainty that occurs when learning from limited data. During meta-training, PLATIPUS uses a structured variational inference approximation during gradient descent to train the initial model and task-specific model parameter distributions. For meta-testing, the initial model parameters are sampled from the distribution and the task-specific parameters are computed by performing maximum a posteriori (MAP) inference on the training set. Bayesian MAML is another probabilistic meta-learning framework that builds upon MAML by incorporating a Bayesian method called Bayesian Fast Adaptation (BFA) to quickly and efficiently update the gradient for a given task by collecting samples from the prior task distribution \cite{yoon_bayesian_2018}. Additionally, they use a novel meta-loss called Chaser loss to prevent overfitting at the meta-level.

\par Some MAML extensions developed methods to improve the alignment of the training task distributions with the test task distribution. Multimodal MAML (MMAML) is able to tackle the tendency to seek a single, shared initialization across the entire task distribution of learning models \cite{vuorio2019multimodal}. It extends MAML by providing it the ability to recognize the mode of tasks sourced from a multimodal task distribution and to quickly adapt through gradient updates. MMAML is designed to adjust its meta-learned prior parameters based on the identified mode, facilitating more efficient and rapid adaptation. Robust MAML (RMAML) incorporates an adaptive learning scheme and a Prioritization Task Buffer (PTB) to increase the training process scalability and mitigate issues arising from distribution mismatch \cite{nguyen2021robust}. To automatically identify the optimal learning rate, RMAML employs gradient-based hyperparameter optimization. It leverages the PTB to incrementally align the training task distribution with the testing task distribution throughout training the model.

\par There has also been work done to extend MAML to other machine learning settings. Follow the meta-leader (FTML) is an online version of MAML that combines MAML with the online learning algorithm follow the leader (FTL) \cite{finn_online_2019}. In the online setting, the learner receives the tasks in sequential order. At each round, the meta-learner determines the optimal model parameters for the specific task presented. The learner's overall goal is to minimize total regret over all tasks.

\subsection{Metric-based Models}
Metric-based meta-learning models generate weights through kernel functions by measuring some distance between two samples. Although the definition of distance varies among metrics, they are all designed to effectively represent the relationship between the feature inputs in the given metric space.

\subsubsection{Siamese neural networks}
\par Siamese neural networks (SNNs) consist of two identical networks that are connected with a shared layer of weights $W$. Through this shared layer, the models are able to share information with one another. The loss function is a contrastive loss given by:

\begin{equation}
    L_{C}(W) = \sum^{N}_{i=1} L(W, (y, x_{1}, x_{2})^{i})
\end{equation}

where the loss between for two inputs $L(W, (y, x_{1}, x_{2})^{i})$ is:

\begin{equation}
    \begin{aligned}
        L(W, (y, x_{1}, x_{2})^{i}) = (1-y) L_{S} (s(x_{1}, x_{2})^{i}) \\
        + \: yL_{D}s(x_{1}, x_{2})^{i})
    \end{aligned}
\end{equation}

and the similarity between two inputs $s(x_{1}, x_{2})$ is: 

\begin{equation}
    s(x_{1}, x_{2}) = ||f_{W}(x_{1}) - f_{W}(x_{2})||
\end{equation}

Note that $L_{S}$ and $L_{D}$ represent the partial loss functions for the similar and dissimilar input pair respectively. Also note that as a result of the shared weights, the loss function used for backpropagation, is additive. This allows the models to learn the similarities and differences between the inputs of the two networks. 
\par A convolutional SNN is similar to an SNN that combines two identical networks into a single model by connecting them with a shared layer \cite{koch2015siamese}; however, these two networks are specifically CNNs. For any given two inputs to these networks $x_{1}$ and $x_{2}$ the label that represents the similarity between them is represented by $\mathbf{y}(x_{1}, x_{2})$. For example, if $x_{1}$ and $x_{2}$ are from the same class, then $\mathbf{y}(x_{1}, x_{2}) = 1$. Otherwise, $\mathbf{y}(x_{1}, x_{2}) = 0$. Additionally, suppose we have the predicted similarity label $\mathbf{o}(x_{1}, x_{2})$, regularization strength $\mathbf{\lambda}$, and network weights $\mathbf{w}$. The loss function is then given by:

\begin{equation}
    \begin{aligned}
        L(x_{1}, x_{2}) = \mathbf{y}(x_{1}, x_{2})\: \mathrm{log} \: \mathbf{o}(x_{1}, x_{2}) \\
        + \: (1 - \mathbf{y}(x_{1}, x_{2}))\: \mathrm{log} \: (1 - \mathbf{o}(x_{1}, x_{2})) + \mathbf{\lambda}^{T}|\mathbf{w}|^{2} 
    \end{aligned}
\end{equation}

It can be seen that as a result of the shared weights, the loss function used for backpropagation, is additive. This allows the models to learn the similarities and differences between the inputs of the two networks.

\subsubsection{Matching Networks}
\par Matching networks use a small support set of data to make informed class label predictions on unseen examples \cite{vinyals2016matching}. The networks use an attention mechanism to determine the similarities between the unseen data and the data from the support set. Suppose we have $x_{u}$, an unseen example, and $x_{i}$, an example from the support set $S = \{(x_{i},y_{i})\}^{m}_{i=1}$. Additionally, $u(x_{u})$ and $v(x_{i})$ are the embeddings produced by the neural networks $u$ and $v$ and $d$ is the cosine distance between these embeddings. The attention mechanism can be represented as:

\begin{equation}
    a(x_{u}, x_{i}) = \frac{e^{d(u(x_{u}),v(x_{i}))}}{\sum^{m}_{j=1} e^{d(u(x_{u}), v(x_{j}))}}
\end{equation}

In this manner, if the embedding of an unlabeled example is similar to a particular class in the support set, then the attention mechanism will award that class a higher weight.

\subsubsection{Prototypical Networks}
Prototypical networks (ProtoNets) first create a prototype representation for each class $\mathbf{c}_{j}$ ($j = \{1,2,\ldots, m\}$) by taking the mean embedding of all the points belonging to that class $S_{j} = \{(x_{i},y_{i})\}^{k}_{i=1}$ \cite{snell_prototypical_2017}. Given a neural network $g$, this mean embedding can be represented by the equation:

\begin{equation}
    \mathbf{c}_{j} = \frac{1}{|S_{j}|}\sum_{(x_{i}, y_{i}) \in S_{j}} g(x_{i})
\end{equation}

By using these class mean embeddings in conjunction with using a distance function $d$, a class probability distribution is created for each data point $x$. This distribution is represented by:

\begin{equation}
    p(y = j \: | \: x) = \frac{e^{-d(g(x),\mathbf{c}_{j})}}{\sum^{m}_{i=1}e^{-d(g(x),\mathbf{c}_{i})}}
\end{equation}

ProtoNets are different from matching networks in few-shot cases but equivalent in the one-shot case. Whereas matching networks use a weighted nearest neighbor classifier for a particular support set, ProtoNets use a linear classifier when using the Euclidean distance as the distance metric.

\subsubsection{Relation Networks}
In relation networks (RNs), the goal is to learn class embeddings in order to be able to transfer this knowledge when predicting classes for unseen test examples \cite{sung_learning_2018}. During meta-training, the RN learns a deep distance metric that is utilized to compute relation scores among a limited number of images in the query and sample sets for a given episode. Given a relation module $r_{\theta}$, an embedding module $h_{\omega}$, and a feature map concatenation function $\mathbf{C}$, the relation score $s_{i,j}$ is represented by:

\begin{equation}
    s_{i,j} = r_{\theta}(\mathbf{C}(h_{\omega}(x_{i}),h_{\omega}(x_{j}))), \quad i = 1,2,\ldots,N
\end{equation}

The parameters $\theta$ and $\omega$ are then optimized using the MSE objective function

\begin{equation}
    \theta, \omega \leftarrow \underset{\theta, \omega}{\mathrm{argmin}}\sum^{k}_{i=1}\sum^{n}_{j=1}(s_{i,j} - \mathbf{1}(y_{i}==y_{j}))^{2}
\end{equation}

Training in this manner replicates the few-shot learning setup that is encountered during testing. After training, the RN can classify images from unseen classes by calculating relation scores between images from the unseen class and query images without the need for updating model weights.

\subsection{Model-based Models}
In model-based meta-learning, a system is developed to ensure swift parameterization for model generalization through integration, analysis, simulation, and synthesis. The system gives the ability to update model parameters quickly and efficiently either through its own model architecture or an external meta-learner.

\subsubsection{Memory Augmented Neural Networks}
Santoro et al.\cite{santoro_meta-learning_2016} proposed memory-augmented neural networks (MANN). A commonly used form of the MANN is the neural Turing machine (NTM). The two major components of NTMs are the controller and the external memory module. The controller produces a key from the input data that can be used to either store the data or retrieve the data from the external memory. A memory $m_{i}$ from episode $i$ is retrieved from the memory matrix $\mathbf{M}_{i}$ using the equation:

\begin{equation}
    m_{i} \leftarrow \sum_{j} w^{r}_{i}(j)\mathbf{M}_{i}(j)
\end{equation}

where $j$ is a particular row in $\mathbf{M}_{i}$ and the read-weight vector $w^{r}_{i}(j)$ is given by the equation
\begin{equation}
    w^{r}_{i}(j) \leftarrow \frac{e^{K(\mathbf{k}_{t}, \mathbf{M}_{i}(j))}}{\sum_{j}e^{K(\mathbf{k}_{i}, \mathbf{M}_{i}(j))}}
\end{equation} 

The cosine similarity measure used to access $\mathbf{M}_{i}$ is given by:
\begin{equation}
    K(\mathbf{k}_{i}, \mathbf{M}_{i}(j)) = \frac{\mathbf{k}_{i} \cdot \mathbf{M}_{i}(j)}{||\mathbf{k}_{i}||\:||\mathbf{M}_{i}(j)||}
\end{equation}

During training, the NTM slowly learns which data is the most relevant and should be stored in external memory using the least recently used (LRU) technique \cite{kumar2016overview}. The equations for the usage weights $\mathbf{w}^{w}_{i}$ and least used weights ${w}^{l}_{i}(j)$ are given by:

\begin{equation}
    \mathbf{w}^{u}_{i} \leftarrow \gamma \mathbf{w}^{u}_{i-1} + \mathbf{w}^{r}_{i} + \mathbf{w}^{w}_{i}
\end{equation}

\begin{equation}
    {w}^{l}_{i}(j) = 
    \begin{cases}
        0 & \text{if } {w}^{u}_{i}(j) > z(\mathbf{w}^{u}_{i},n) \\
        1 & \text{if } {w}^{u}_{i}(j) \leq z(\mathbf{w}^{u}_{i},n)
    \end{cases}
\end{equation}

where $z(\mathbf{w}^{u}_{i},n)$ is the \textit{n}th smallest element in $\mathbf{w}^{u}_{i}$. From these equations, it can be seen that rarely used or old data is removed from the external memory over time, while recently stored or frequently used data is retained in the external memory. When tested on an unseen task, the NTM can then access the external memory to retrieve the most relevant information in order to quickly learn the task.

\subsubsection{Meta Networks}
A meta network (MetaNet) is essentially a MANN with three components: a meta learner $v$, a base learner $b$, and an external memory module $M$ \cite{munkhdalai_meta_2017}. The meta learner works in the task-agnostic meta space, and the base learner works in the input task space. The meta space allows the meta learner to promote continual learning and attain meta knowledge across different tasks. The base learner analyzes the input task and then gives the meta learner feedback through meta information to explain its status in the current task space. The meta learner uses this meta information to quickly parameterize itself and the base learner so the MetaNet model can quickly recognize new input task concepts. As a result, the MetaNet training weights are updated at different frequencies. The slow weights of the meta learner and the base learner are updated through a learning algorithm, the fast weights of the meta learner are updated for each task, and the fast weights of the base learner are updated for each input. The external memory also assists in quick learning and generalization. The algorithm is summarized in Algorithm \ref{alg:metanet}.

\begin{algorithm}
    \caption{MetaNet}
    \label{alg:metanet}
    \begin{algorithmic}
        \REQUIRE Support set $\{x^{s}_{i},y^{s}_{i}\}^{N}_{i=1}$ and training set $\{x^{t}_{i},y^{t}_{i}\}^{L}_{i=1}$
        \REQUIRE Base learner \textit{b}, dynamic representation learning function \textit{v}, fast weight generation functions $g$ and $f$, and slow weights $\theta = \{W,Q,Z,G\}$
        \REQUIRE Layer augmentation scheme
        \STATE Sample \textit{K} examples from support set
        \FOR{$i = 1, K$}
            \STATE $\mathcal{L}_{i} \leftarrow loss_{repr}(v(Q^{s},x^{s}_{i}),y^{s}_{i})$
            \STATE $\nabla \leftarrow \nabla_{Q^{s}}\mathcal{L}_{i}$
        \ENDFOR
        \STATE $Q^{f} = f(G,\{\nabla\}^{T}_{i=1})$
        \FOR{$i = 1, N$}
            \STATE $\mathcal{L}_{i} \leftarrow loss_{task}(b(W^{s},x^{s}_{i}),y^{s}_{i})$
            \STATE $\nabla_{i} \leftarrow \nabla_{W^{s}}\mathcal{L}_{i}$
            \STATE $W^{f}_{i} \leftarrow g(Z,\nabla_{i})$
            \STATE Store $W^{f}_{i}$ in $i^{\mathrm{th}}$ position of memory \textit{M}
            \STATE $r'_{i} = v(Q^{s},Q^{f},x^{s}_{i})$
            \STATE Store $r'_{i}$ in $i^{\mathrm{th}}$ position of index memory \textit{R}
        \ENDFOR
        \STATE $\mathcal{L}_{train} = 0$
        \FOR{$i = 1, L$}
            \STATE $r_{i} = v(Q^{s},Q^{f},x^{t}_{i})$
            \STATE $a_{i} = attention(R,r_{i})$
            \STATE $W^{f}_{i} = softmax(a_{i})^{\top}M$
            \STATE $\mathcal{L}_{train} \leftarrow \mathcal{L}_{train} + loss_{task}(b(W^{s}, W^{f}_{i}, x^{t}_{i}),y^{t}_{i})$ \\
            (\textbf{Note:} The base learner can also use $r_{i}$ instead of $x^{t}_{i}$ as input)
        \ENDFOR
        \STATE Update $\theta$ using $\nabla_{\theta}\mathcal{L}_{train}$
    \end{algorithmic}
\end{algorithm}

\subsubsection{Simple Neural Attentive Meta-Learner}
Unlike MANNs and MetaNets, the Simple Neural Attentive Meta-Learner (SNAIL) does not rely on an external memory architecture \cite{Mishra2017}. Instead, it uses temporal convolutional networks to enable memory access and attention mechanisms to locate specific chunks of memory. SNAIL consists of a DenseBlock, the TCBlock, and the AttentionBlock. The DenseBlock uses a convolution on the input data and concatenates the result. The TCBlock consists of a series of DenseBlocks of varying dilation rates. This helps to distill pertinent information from the input data. The AttentionBlock allows SNAIL to focus on the important parts of the prior experience.

\subsubsection{Neural Statistician}
The Neural Statistician \cite{Edwards2016} also does not rely on the use of an external memory. Rather, it develops summary statistics of the datasets using an unsupervised process. These statistics can then be used to make predictions on unseen observations. During training, the neural statistician creates generative models for each dataset. The training process is optimized using a loss function consisting of reconstruction loss, context loss, and latent loss. The reconstruction loss measures how well the model generates an underlying data distribution similar to the input data. The context loss evaluates how well the context corresponds to the prior distribution. The latent loss assesses how well the latent variables are modeled.

\subsubsection{Conditional Neural Processes}
Similar to SNAIL and the Neural Statistician, conditional neural processes (CNPs) \cite{Garnelo2018} also do not use external memory to operate. Instead, they create embeddings of the observations and ground truth labels from the support set and aggregate these embeddings into a single representation. These final representations are then used in conjunction with unseen test observations to form predictions.

\section{Multi/Single-task Learning}
\begin{figure}
  \centering
  \includegraphics[width=0.45\textwidth]{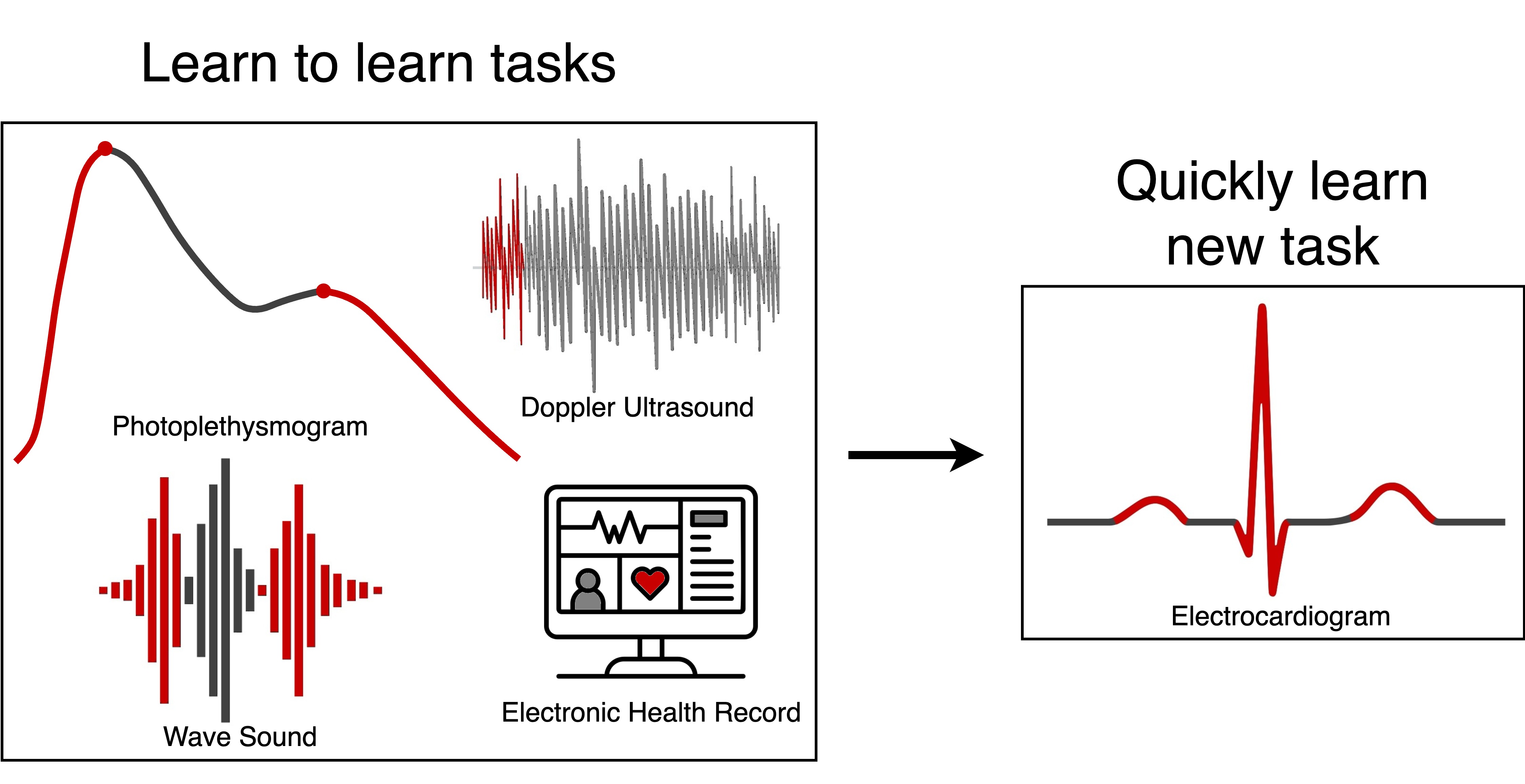}
  \caption{Rapid adaptation example of a meta-learning model to a new task by leveraging experience from various prior data sources in healthcare.}
  \label{fig:multi}
\end{figure}

Multi-task learning involves simultaneously training a model on several related tasks, thereby leveraging shared information across these tasks to augment overall performance. This approach enables the model to generalize better to unseen tasks by capturing common patterns and structures among different tasks. Single-task learning centers around training a model specifically for one task, often yielding a model that is finely tuned to that particular task. Within the realm of meta-learning, both multi-task and single-task learning methodologies contribute to developing more adaptable and robust models. Of note, it can bolster the model's capacity to learn from sparse data and generalize to new tasks, addressing significant challenges in a variety of applications such as clinical risk prediction, disease detection, and drug discovery. Figure \ref{fig:multi} shows the ability of a meta-learner to gather insights from a variety of data types and tasks to quickly adapt to a new task. In the context of this paper, all non-image multi-task and single-task studies that used a meta-learning technique were analyzed in this section.

\subsection{Clinical risk prediction and diagnosis}
Clinical risk prediction models (CRPMs) aim to estimate the probability of specific clinical outcomes using a variety of patient-specific parameters to strengthen medical decision-making. Over the past decade, there has been a surge in the development of CRPMs across all areas of healthcare. Predictive modeling for clinical risk can be applied to various data types, such as electronic health records (EHR), electroencephalograms (EEG), and electrocardiograms (ECG). It can also be employed in different applications, such as disease onset, condition exacerbation, and mortality prediction \cite{goldstein2017opportunities,rafiei2021ssp}. The adoption pathway of automated CRPMs generally requires access to rich data, regulatory approval, and seamless integration into clinicians' workflows \cite{sharma2021clinical}. CRPM development typically faces the limited availability of data and the high cost of acquiring additional data. In addition, the data available for CRPMs is often characterized by noise, sparsity, irregularity, and sequentially \cite{wang2012towards}. Consequently, there is a critical need for reliable CRPMs that can be optimally developed using limited data. Based on the clinical input data used in CRPMs, they can be classified into three primary categories: EHR, EEG, and ECG. The challenges and meta-learning solutions for each CRPM category are discussed below.

\subsubsection{Electronic Health Records}
An EHR is a digital compilation of a patient's medical history, encompassing clinical data from a specific healthcare provider. This data can include demographics, medications, medical issues, vital signs, laboratory results, past medical history, immunizations, and radiology reports \cite{rafiei2024robust}. EHRs can streamline a clinician's workflow by automating access to information and supporting care-related activities through various interfaces such as quality management, evidence-based decision support, and outcome reporting \cite{kohli2016electronic, rafiei2024improving}. The frequency of data collection for EHRs varies among hospitals and cases. This leads to a disparity in the volume of records available for different medical conditions. Put simply, the amount of collected data for patients afflicted with a certain disease can vastly outweigh the data available for patients dealing with another medical condition. To exploit the labeled EHRs from a high-resource domain, Zhang et al. \cite{zhang2019metapred} developed a meta-learning framework, MetaPred, for clinical risk prediction. This framework leverages meta-learning-based models to transfer knowledge from source domain data (data from related diseases with a sufficient number of labels) to a resource-limited target domain. The risk prediction model in MetaPred can be configured as either a convolutional neural network (CNN) or LTSM model in which an optimization-based adaptation using gradient update loops was employed. In addition, an objective-level adaptation to compensate for the fast adaptation at the optimization level was proposed. This strategy improves the objective by integrating supervision from available source domains. The effectiveness of the MetaPred learning procedure was evaluated on five cognitive diseases EHR datasets for risk prediction tasks. 

\par Likewise, Tan et al. \cite{tan2022metacare++} proposed MetaCare++ as a cold-start diagnosis prediction model for infrequent patients (a cohort of patients with only a few recorded visits) and rare diseases (diseases affecting a small percentage of the population), using EHR data. This clinical meta-learner leverages an autoencoder structure for diagnosis prediction, taking into account the effects of disease progression on patients over time. The autoencoder utilized continuous normalizing flow (CNF) \cite{mathieu2020riemannian}, which alleviates the meta-learning framework's inference gaps and enhances its performance by modeling highly complex distributions. To model the sophisticated associations among rare diseases, they captured hierarchical and syndromic associations between diseases using domain knowledge and actual diagnosis subtyping data. By integrating these two patient-based and disease-based approaches, MetaCare++ was formed and evaluated on real-world benchmark datasets.

\par To address the security challenges of sharing sensitive EHRs, Thakur et al. \cite{thakur2021dynamic} developed a federated meta-learning framework. The framework utilized a dynamic neural graph learning (NGL) \cite{bui2018neural} method to incorporate unlabeled EHRs in the learning process. This semi-supervised meta-learning approach enables the model to perform multiple tasks by improving generalizability. The model's performance on the MIMIC-III dataset demonstrated promising results in overcoming the constraints imposed by limited supervision and data decentralization.

\par Patients admitted to the intensive care unit (ICU) setting may have varying lengths of stay, frequencies of vital sign collection, and diverse types and numbers of laboratory tests. The heterogeneity in ICU data arises from the differences in risk and the severity of patients' conditions. Therefore, implementing a unique model for every ICU patient and condition may not be optimal. Zhang et al. \cite{zhang2021dynehr} developed DynEHR, a model designed to adapt to various ICU lengths of stay. This approach employs an LSTM model on the time-series data and applies MAML on the top of the LSTM layers. The performance of DynEHR was evaluated on the MIMIC-III dataset for different tasks. Similarly, An et al. \cite{an2021prediction} proposed a dual adaptive sequential network (DASNet) to predict treatment medicines with the aim of capturing various correlations of heterogeneous EHR sequences. To achieve this, they considered two decomposed adaptive LSTM networks in which an attentive meta-learning network dynamically generated the weight parameters. This approach consisted of a meta-LSTM and two history attention networks to make the historical meta-knowledge more informative. The model was evaluated on 11 extracted subsets of the MIMIC-III dataset.

\subsubsection{Electroencephalogram}
An EEG is a recording of dynamic brain activity that is obtained using electrodes attached to the scalp \cite{cohen2017does}. The electrodes detect minute electrical signals originating from the activity of neurons. These neurons, which are always active, communicate via electrical impulses. The activity for communications can be measured by the electrodes, showing wavy lines on an EEG record \cite{kumar2012analysis}. The data provided can help physicians detect or predict a wide range of diseases, disorders, and difficulties \cite{siuly2016significance,siuly2016significance}. Due to many challenges in the manual interpretation of EEG records \cite{tatum2021handbook,tatum2012eeg}, different ML and DL algorithms have been developed to help in translating EEG data \cite{ruffini2019deep,rafiei2022automated,koleck2019natural}. However, most of these models are dataset-specified and achieve significant performance on only one or a limited number of datasets. To this end, Liu et al. \cite{liu2022side} introduced the Side-aware Meta-learning (SMeta) model for cross-dataset tinnitus diagnosis. SMeta used an autoencoder structure for the classification of participants while it reconstructed raw signals to extract tinnitus-specific information from input EEG data. This model employed a down-sampling policy to align the number of time points for different EEG datasets. Hence, SMeta could efficiently control and classify participants of divergent ages and genders using different data collection processes.

\par Personalization of emotion prediction and classification of patients are other areas that can be facilitated by applying meta-learning on brain signals. Since EEGs differ for each individual, numerous methods have been employed to train an independent emotion induction model using a single participant's EEG data \cite{li2018exploring,miyamoto2020music,ehrlich2019closed}. As this approach was not generalizable and did not represent encouraging outcomes, Miyamoto et al. \cite{miyamoto2021meta} proposed a meta-CNN model for emotion prediction, which can distinguish individuals from multiple participants when listening to music. They found that by considering each participant as an independent task, the training process could effectively accommodate the individual nature of EEG data. Likewise, Wang et al. \cite{wang2022inference} introduced the Anes-MetaNet classification model to mitigate the individual differences among anesthesia patients in an office-based environment. They employed a meta-CNN along with an LSTM network to outperform ordinary DL models. In addition, the combination of feature extraction and meta-transfer learning techniques holds promise for overcoming the individual difference issue in cross-subject EEG emotion recognition tasks \cite{li2022cross,tang2022deep,duan2020meta}.

\subsubsection{Electrocardiogram}
An ECG is a simple, fast test to evaluate the heart by measuring its electrical activity \cite{berkaya2018survey}. Small electrodes are located at certain spots on the body, and a machine records the heart signals to be used for detecting heart-related problems. Doctors can diagnose a wide range of abnormalities by analyzing ECGs, such as arrhythmias, heart attacks, and cardiomyopathy \cite{kusumoto2020ecg}. Although interpreting ECG data is more straightforward than EEG for physicians, achieving a high accuracy is still challenging \cite{cook2020accuracy}. Despite ML algorithms achieving significant success in analyzing ECGs, they face challenges in managing the diversity of data at the group level (cohort of patients with a few recorded visits) and individual level (stage differences of a single subject). To meet this challenge, Zhang et al. \cite{zhang2022metava} proposed a two-step meta-learning procedure known as MetaVA to detect ventricular arrhythmias (VA). First, they combined MAML with a curriculum learning (CL) selector strategy to pre-train a deep CNN model. CL-selector is a training strategy to afford an easy-to-hard learning scheme for MAML \cite{bengio2009curriculum}. This approach adopts the human learning process through which learning starts from basic to advanced concepts. As a result, they measured the complexity of every meta-tasks and ordered from easy to difficult for training the model. Second, they adopted a pre-fine-tuning method to solve individual-level diversity as well as a conventional fine-tuning approach to improve the performance of the developed model. Evaluation of MetaVA on three large datasets demonstrated that this model could quickly adapt to new individuals with acceptable performance metrics. Meqdad et al. \cite{meqdad2022meta} developed an interpretable meta-structural learning model to classify seven types of arrhythmia for different patients within distinct sessions. To do so, they considered a compound loss function incorporating space label error with the GUMBEL-SOFTMAX distribution and structural feature extraction error in a CNN model with thirteen convolutional layers. Additionally, a genetic programming (GP) algorithm was employed to encode the CNN model in evolutionary trees, making it more interpretable. Lin et al. \cite{lin2018cardiac} proposed a meta-learning-based classifier for screening cardiac arrhythmias, dealing with the incremental learning of newly available training patterns. Their incremental learning scheme included a primary multilayer network and seven agent networks. The multilayer network was a generalized regression neural network (GRNN), fine-tuned by the particle swarm optimization (PSO) algorithm. The agents were employed to progressively enhance the optimization by leveraging the previous learning experiences for adaptive applications. This approach led to accurate recognition and computational efficiency for real-time ECG automated screening.

\subsubsection{Wearable physiological sensing}
The use of wearable devices capable of measuring vital signs has become increasingly popular. Today, many people wear smart watches and bands in their daily lives that monitor several physiological signals, such as heart rate, blood pressure, and temperature. Despite the growing development of healthcare-related DL algorithms with the aim of real-world implementation on these devices, a limited number of automated models have been put into practice \cite{iqbal2021advances}. The reason is that in addition to considering individual and group level diversity for the data from wearable devices, sensor and hardware-configuration level differences must also be addressed. These considerations are necessary for a successful real-world implementation of a clinical risk monitoring algorithm. To tackle this problem, Hao et al. \cite{hao2021invariant} developed an invariant feature learning framework (IFLF) based on meta-learning to decipher common information across subjects and devices. Their model encompassed four CNN layers and two LSTM layers to handle various sources of domain shifts. The performance of the proposed method was assessed in extensive experiments over several datasets with different subjects, sensors, and hardware configurations. Similarly, Nithin et al. \cite{nithin2021sensor} aimed to address mobility status diversity in elderly individuals using meta-learning for human activity recognition (HAR) tasks.

\par Jia et al. \cite{jia2022personalized} established a meta-learning method for patient-specific detection tasks on resource-constrained Internet of Things devices. They proposed a new patient-wise training task formatting strategy to meet the training sample mixture challenge in conventional meta-learning models. Also, inner and outer loop optimizations were developed for further improvement of the generalization of the meta-model initialization. This method was evaluated on three different health monitoring tasks: ventricular fibrillation (VF) detection, atrial fibrillation (AF) detection, and HAR. Filosa et al. \cite{filosa2022meta} proposed a meta-LSTM model to provide an accurate and subject-specific model for the prediction of respiratory flow. Akbari et al. \cite{akbari2021meta} introduced a modality translation framework for the personalized translation of bio-impedance (Bio-Z) to an ECG signal using a meta-CNN model. The framework used a single neural network architecture to learn personalized models for each user from a multi-user dataset by automatically adjusting the network's parameters. It was able to adapt the models to new users with minimal samples of new data. The performance was evaluated for Bio-Z to ECG translation using real data from wearable Bio-Z patches.

\subsection{Automated Medical Detection}
Automated medical detection refers to the application of learning algorithms in healthcare. Medical detection is a complex task and often necessitates ample investigation of clinical situations. This process can place a substantial cognitive load on clinicians, as they strive to extract meaningful information from intricate and varied clinical reports. Also, ML models are heavily dependent on a substantial quantity of labeled samples to infer and detect possible conditions. Although the scope of this section can be broader, we have discussed the sub-categories here that have not been included in previous sections.

\subsubsection{Diabetes}
Diabetes mellitus, more commonly known as diabetes, refers to a serious health condition that affects how the body uses blood glucose. The International Diabetes Federation (IDF) estimates that approximately 463 million adults worldwide are afflicted by diabetes, resulting in roughly four million fatalities and a global healthcare expenditure of \$727 billion \cite{saeedi2019global}. The worldwide prevalence of diabetes is anticipated to rise to 454 million by 2030 and 548 million by 2045 \cite{saeedi2019global}. In addition, patients with diabetes are associated with a longer hospital length of stay (LOS) and higher frequency of admission \cite{cheng2019costs,comino2015impact}. Given its widespread occurrence, substantial economic impact, and societal implications, the development of innovative and automated approaches to address diabetes is of paramount importance.

\par A notable concern of the diabetic population is the demand for patient-specific and personalized approaches to blood glucose (BG) regulation \cite{yu2021reinforcement}. The variability in lifestyle factors, such as exercise, diet, stress levels, and illness, combined with the complex nature of the glucose-insulin regulation system, highlights the necessity for more advanced adaptive algorithms for BG management. In this regard, Zhu et al. \cite{zhu2022personalized} proposed a fast-adaptive model based on evidential DL and meta-learning for BG prediction with limited data. Nemat et al. \cite{nemat2022blood} developed an ensemble model for predicting BG 60 minutes in advance. The outputs of the ensemble model were then fused using three different meta-learning-based approaches. In the first approach, the outputs of the learners of the ensemble model were concatenated and fed to a linear model as the meta-learner. In the second approach, the learners' outputs were considered as different variables and passed onto a Meta-LSTM. The third approach involved considering the learners' outputs as different subsequences to be passed onto the convolutional LSTM (ConvLSTM) meta-learner.

\subsubsection{Gene}
The number of genetic testing is proliferating in clinical care to diagnose, prevent, manage, and treat an extensive diversity of diseases, syndromes, and disorders. Over the years, genetic testing has evolved from single-gene analysis to the feasibility of surveying the entire genome of a patient for disease-causing variants \cite{clark2018meta,bertoli2021successful}. Nevertheless, despite whole-exome sequencing (WES) and whole-genome sequencing (WGS) being powerful tools to determine which identified variants are more likely to underlie the disease phenotype and damage gene function, less than 50\% of Mendelian disorders resolve after sequencing of affected families \cite{chong2015genetic,eilbeck2017settling}. This performance is even more inadequate for multifactorial and complex diseases \cite{ahmed2021genetic}. Studies revealed that a quarter of individuals who have a rare disease wait between 5 to 30 years and appoint three doctors at least to get a diagnosis, with a wrong initial diagnosis in at least 40\% cases \cite{robinson2017computational}. There are several contributing factors why it is perplexing to discern the identity of a causal gene: new genes for known conditions, unprecedented disease phenotypes, and a large number of undiscovered disease-causing variants. A typical exome contains up to 140,000 variants, with around 200 or more never-before-seen missense variants, while manual processes for clinical analysis and reporting variants are incredibly slow \cite{cooper2011needles}. Therefore, employing automated systems such as ML models is necessary to boost the analysis of genome data. As the number of available sequenced data is still limited, every genome contains a massive amount of information, and the generalization capability of a developed model is notable, meta-learning can be a powerful tool for genetic-related analysis.

\par Reproducibility of results can greatly influence the adoption of novel technologies in analyzing gene expression. Stiglic et al. \cite{stiglic2010gene} proposed a meta-learning-based approach to assess the consistency of gene ranking results and compare results obtained from two types of genomic data, next-generation sequencing (NGS) and microarrays. The developed similarity-based decision tree model could extract knowledge from multiple bootstrapped gene set enrichment analyses. This study ultimately enables researchers to make more informed decisions when choosing between NGS and microarray technologies for gene expression. To address the limited number of samples and high feature count for genomic survival analysis, Qiu et al. \cite{qiu2020meta} presented a multi-layer perceptron (MLP) model followed by a Cox loss as the meta-learner. The model was able to prioritize genes based on their contribution to survival prediction. They evaluated this approach to real-world cancer datasets in terms of robustness and generalization performance using statistical methods.  Zhou et al. \cite{zhou2022annotating} employed a meta-learning approach to enhance the prediction of transcription start sites (TSSs) across multiple cell types. Their meta-CNN framework allows the model to adapt to new cell types quickly and efficiently, improving the generalization of TSS annotations with handling the complex, dynamic nature of gene regulation and transcription. Jiang et al. \cite{jiang2022spectral} proposed SiaClust, a meta-learning strategy designed to address the issue of cell clustering due to the presence of imbalance and high-dimensionality in datasets and enables effective cell classification even with limited scRNA-seq data. SiaClust used convolutional SNN and improved the affinity matrix. The network learns cell-to-cell similarity, while the improved matrix optimizes graph structure. This model was evaluated on the nine human and mouse scRNA-seq datasets with different cell numbers and types.

\subsubsection{Voice}
Voice technology in healthcare involves processing and analyzing human speech and organ sounds for medical-related applications \cite{latif2020speech}. With the growth of automated voice-controlled systems and sound recorders, gathering data for healthcare applications has become more straightforward, making automated models ready to play a significant role in the diagnosis process. Yet, the insufficient amount of data, cross-lingual settings, and domain shifts are the bottlenecks of this type of data to developing generalized ML and DL models. Tackling these challenges with meta-learning, recently, has attracted the attention of researchers to develop more reliable models. Chopra et al. \cite{chopra2021meta} proposed a cross-lingual speech emotion recognition (SER) based on meta-learning called MetaSER for low-resource languages. They developed an LSTM-based model aimed at learning language-independent emotion-specific features. Koluguri et al. \cite{koluguri2020meta} presented a meta-learning model to perform child-adult speaker classification within the autism spectrum disorder (ASD) in spontaneous conversations. This approach employed PNs to address the lack of a sufficient number of balanced training data. Ditthapron et al. \cite{ditthapron2021learning} explored a Meta-CNN model to mitigate overfitting of traumatic brain injury (TBI) detection from speech with limited data. Barhoush et al. \cite{barhoush2022localization} presented a convolutional RNN (CRNN) model for 3D localization and enhancement of the speech of a main speaker in multispeaker, noisy, and reverberant hospital environments. They set up a meta-learning method to ensure that the speech enhancement approach is adaptive to various conditions and environments.

\subsection{Drug Development}
\subsubsection{Discovery and Interaction}
Drug discovery is the process of discovering novel candidate medications \cite{atanasov2021natural}. This costly and time-consuming procedure typically takes a decade to complete and has a low success rate \cite{kola2004can}. To facilitate the drug development cycle, numerous computational methods, such as ML and DL models, are proposed to virtually design, simulate, and test molecules \cite{dara2021machine}. In recent years, different meta-learning approaches have been applied in various aspects of drug discovery and development. Olier et al. \cite{olier2018meta} investigated the utility of meta-learning for quantitative structure-activity relationship (QSAR) analysis. A standard QSAR learning problem aims to learn a predictive function of molecular representation to activity from a given target (typically a protein) and a set of chemical compounds with associated biological activities \cite{kwon2019comprehensive}. They trained a random forest meta-learner on a large number of datasets to select an ML algorithm expected to perform best on given meta-features. These QSAR-specific meta-features involved measurable characteristics of datasets, such as normalized entropy of the features and drug target properties. Wang et al. \cite{wang2021meta} proposed the Meta-MO approach for molecular optimization (MO) to improve the pharmaceutical properties of a starting molecule. Meta-MO leveraged bioactivity data in distinct protein targets to learn a model from resource-rich targets that can be generalized for other low-resource MO tasks. For this aim, a graph-enhanced transformer (GET) model was developed and optimized to generate optimized molecules from a source molecule input.

\begin{table*}[]
\centering
\small
\caption{Summary of meta-learning applications in multi/single-task learning.}
\begin{tblr}{
colspec={p{1.75cm} p{1.5cm} p{0.65cm} p{1.7cm} p{1.75cm} p{1.5cm} p{5.5cm}},
  cell{2}{1} = {r=11}{},
  cell{2}{2} = {r=5}{},
  cell{7}{2} = {r=2}{},
  cell{9}{2} = {r=3}{},
  cell{13}{1} = {r=7}{},
  cell{14}{2} = {r=3}{},
  cell{17}{2} = {r=3}{},
  vline{2},vline{3},vline{4},vline{5},vline{6},
  hline{1-2,13,20-21} = {-}{},
  hline{3-6,8,10-11,15-16,18-19} = {3-8}{},
  hline{7,9,12,14,17} = {2-8}{},
  vline{7}
}
\hline
\textbf{Domain}                                 & \textbf{Application}                    & \textbf{Ref.}        & \textbf{Strategy}                        & \textbf{Methods}                                         & \textbf{Data Acquisition}                   & \textbf{Highlights or Limits}                                                                                                                     \\
Clinical risk prediction and diagnosis & EHR                            & \cite{zhang2019metapred}     & Optimization-based, Model-based & Meta-CNN, Meta-LSTM Source domain adaptation    & OCTRI                              &Consider an objective level adaption to compensate for the fast adaptation at the optimization level.                                   \\
                                       &                                & \cite{tan2022metacare++}       & Optimization-based, Model-based    & Meta-Autoencoder, Hierarchical subtyping        & MIMIC-III, eICU                   & Propose a predictive model for infrequent patients and rare diseases.                                                                    \\
                                       &                                & \cite{thakur2021dynamic}    & Objective-based                 & Meta-NGL                                       & MIMIC-III                         & Develop a federated meta-learning that can use unlabeled data for the training process.                                                  \\
                                       &                                & \cite{zhang2021dynehr}     & Optimization-based              & Meta-LSTM                                       & MIMIC-III                         &  Propose a dynamic adaption of models for ICU heterogeneity EHRs.                                                                         \\
                                       &                                & \cite{an2021prediction}        & Model-based                     & Attentive MetaNet                               & MIMIC-III                  & Predict treatment medicines for capturing various correlations of heterogeneous EHR sequences.                                           \\
                                       & EEG                            & \cite{liu2022side}       & Optimization-based              & Meta-Autoencoder                                & Two tinnitus datasets         & Introduce a cross-dataset model to achieve a generalized diagnosis of tinnitus.                                                          \\
                                       &                                & \cite{miyamoto2020music}  & Optimization-based              & Meta-CNN                                        & 20 EEG records                    & Develop an emotion induction model to generate music considering the individuality based on predicted emotion.                           \\
                                       & ECG                            & \cite{zhang2022metava}     & Optimization-based              & Meta-CNN, CL-selector                           & MITDB, CUDB, VFDB                  & Combine MAML with a CL selector strategy and considering a pre-fine-tuning method to address group-level and individual-level diversity. \\
                                       &                                & \cite{meqdad2022meta}    & Metric-based                    & CNN tree, GP                                    & Chapman                       & An interpretable meta structural learning for classifying seven types of arrhythmia.                                                     \\
                                       &                                & \cite{lin2018cardiac}       & Model-based, Optimization-based & Meta-GRNN, PSO                                  & MIT-BIH arrhythmia database     & Screen cardiac arrhythmias using an incremental learning strategy.                                                                       \\
                                       & Wearable physiological sensing & \cite{hao2021invariant}       & Metric-based                    & ConvLSTM, customized similarity-based function~ & PAMAP2, USCHAD, WISDM, MobilityAI  & Decipher common information across subjects and devices for HAR tasks.                                                                   \\
Automated medical detection            & Diabetes                       & \cite{zhu2022personalized}       & Optimization-based              & Bidirectional GRU                               & OhioT1DM, ARISES, ABC4D        & Propose a fast-adaptive model for BG prediction with a limited number of data.                                                           \\
                                       
                                       & Gene                           & \cite{qiu2020meta}       & Optimization-based              & MLP, Cox loss                                   & TCGA                               & Conduct a survival analysis on a small sample size that has a high number of features.                                                   \\
                                       &                                & \cite{zhou2022annotating}     & Optimization-based              & Meta-CNN                                        & FANTO, 5 CAGE, ENCODE              & Improve TSS prediction generalization across multiple cell types.                                                        \\
                                       &                                & \cite{jiang2022spectral}     & Metric-based                    & Convolutional SNN, spectral clustering          & 9 scRNA-seq data sets           & Address the issue of imbalanced data across various cell types within scRNA-seq datasets.                                                \\
                                       & Voice                         & \cite{chopra2021meta}    & Optimization-based              & Meta-LSTM                                       & 8 language datasets              & Present a cross-lingual SER model for low-resource languages.                                                                            \\
                                       &                                & \cite{koluguri2020meta}~ & Metric-based                    & ProtoNet                                        & A private dataset               & Address the lack of balanced speech data.                                                                                                \\
                                       &                                & \cite{barhoush2022localization}   & Optimization-based              & CRNN                                            & Simulated and real ICU sounds     & Generalize a speech enhancement model for various hospital conditions and environments.                                                  \\
Drug development                       & Discovery and interaction      & \cite{wang2021meta}      & Optimization-based              & GET                                             & ChEMBL20                           & Propose and optimize a Transformer-based meta-model for MO tasks. \\
\hline
\end{tblr}%

\end{table*}

\subsection{Insights and implications}
The multi/single-task learning section showcases special applications in healthcare by providing innovative solutions to various challenges. In addition to tackling data limitations and handling distribution shifts, these meta-learning approaches lead to rapid cross-dataset generalization, adaptation, and enhanced personalization. This allows models to adapt to individual patients or specific clinical scenarios, accounting for individual physiological differences while maintaining generalizability, capabilities not easily achieved by other ML models. The analysis of the studies reveals that one of the challenges for adopting meta-learning techniques is the potential for negative transfer, where learning multiple tasks simultaneously can lead to interference, reducing the model's performance on some tasks. Techniques like CL and adaptive weighting of task losses offered promising solutions. Besides, there is a notable lack of standardized metrics for comparing performance and other aspects of developed models. Most studies used their specific metrics and reported task-specific results, making comparisons with other approaches difficult. They often did not compare their results with other meta-learning or ML models, nor did they assess other aspects of their models, such as latency, generalizability, and bias. Optimization- and model-based approaches are commonly used for different tasks; however, metric-based methods have the potential for baseline comparisons and assess generalizability. Overall, while the core principles of a specific meta-learning technique can be applied to different tasks and data modalities, the specific implementations may vary due to inherent differences in data characteristics, task-specific requirements, and evaluation assessments.

\section{Many/Few-shot Learning}
\begin{figure}
  \centering
  \includegraphics[width=0.45\textwidth]{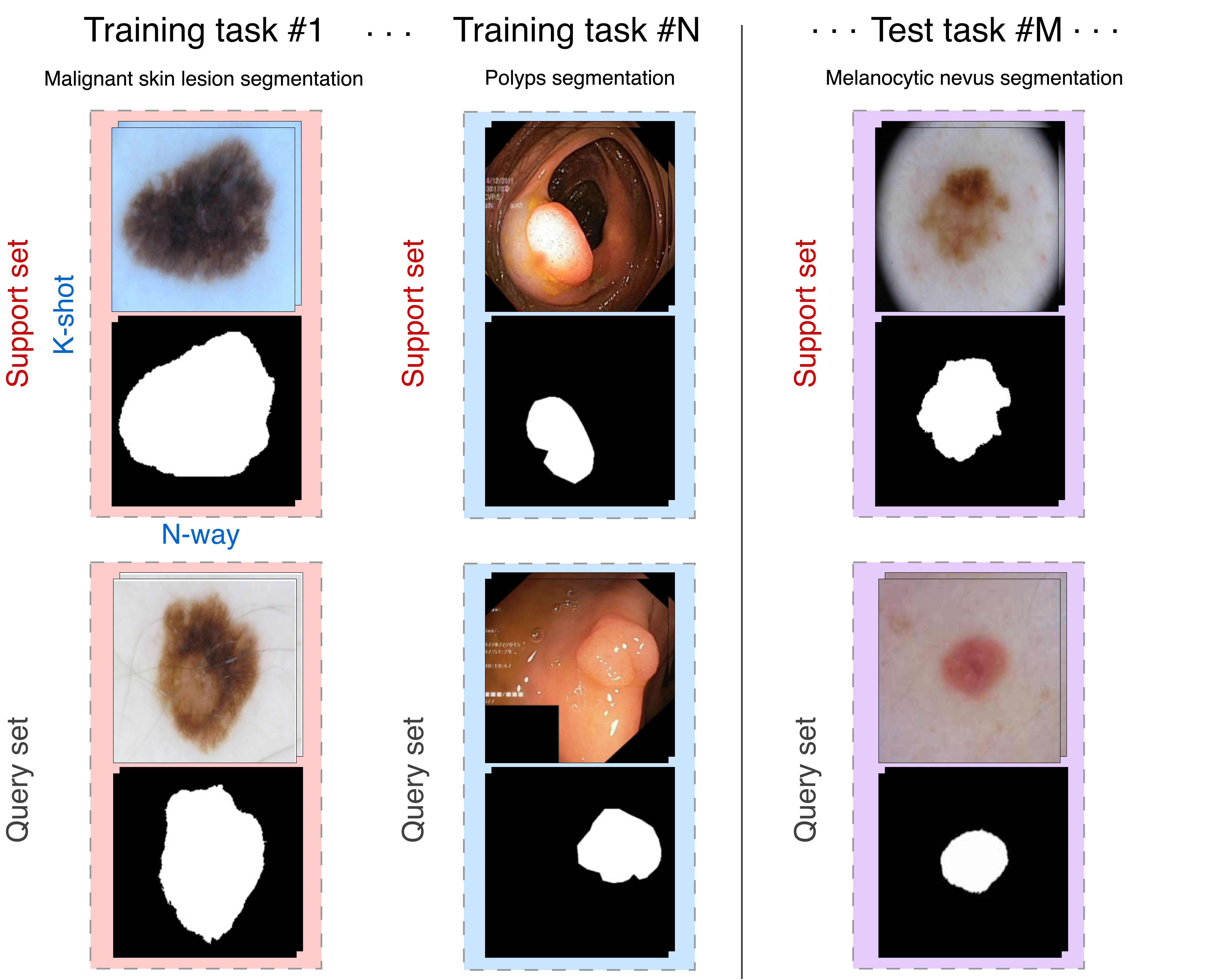}
  \caption{Example of many/few-shot medical image segmentation using meta-learning.}
  \label{fig:few}
\end{figure}

Multi-shot and few-shot learning are primarily designed to aid models in generalizing from a small number of examples. Multi-shot learning refers to situations where the model is trained on multiple instances of each class in a dataset, which helps the model to learn a more robust representation of each class. On the other hand, few-shot learning is specifically tailored for situations where only a handful of examples are available for each class. This approach is particularly useful in settings where collecting large amounts of labeled data is challenging or impractical. Despite the scarcity of data, few-shot learning techniques aim to build models that can still perform well on new, unseen data. Figure \ref{fig:few} illustrates a meta-learning approach for medical image segmentation, where the model is developed using support and query sets for several different tasks with K-shot (number of examples per class) and N-way (number of classes) and then uses prior knowledge to perform segmentation on new tasks.

\par Medical conditions can generally be classified into two primary categories: chronic with comorbidities and acute. This categorization is based on the causes, symptoms, and treatment required for each type of ailment. Chronic conditions are characterized by symptoms that persist for an extended period, typically more than three months, and are generally non-contagious. These conditions often necessitate long-term management strategies to alleviate symptoms and maintain overall health \cite{bernell_use_2016}. Comorbidities refer to the simultaneous presence of two or more medical conditions in a single individual. In contrast, acute illnesses are generally characterized by a rapid onset and a relatively short duration. These conditions often resolve spontaneously or with medical intervention. Unlike chronic conditions, acute illnesses do not typically have lasting effects on the individual's health once resolved.

\par In this section, we focused our analysis solely on studies that applied meta-learning within the context of many/few-shot learning, specifically addressing the two aforementioned medical conditions. The input data for these studies could be either text or images, with no restrictions regarding disease types or tasks performed by the models.

\subsection{Chronic Diseases and Comorbidities}
Chronic and comorbid conditions represent a significant challenge for both individuals and healthcare systems. The combination of multiple medical conditions can complicate treatment and management, increasing the risk of negative health outcomes and reducing the quality of life. As the prevalence of chronic and comorbid conditions continues to rise, particularly in aging populations, it becomes increasingly important to develop effective strategies for early detection, prevention, management, and treatment \cite{vos_global_2020, linden2022prevalence}. These strategies should not only focus on addressing individual chronic conditions but also consider the complex interplay between comorbid diseases.

\subsubsection{Cancer}
\par Cancer, with over 150 distinct forms, has seen a rise in cases due to factors such as increased tobacco use, unhealthy lifestyles, poor dietary choices, and genetic disorders. In 2018 and 2020, the number of cancer incidents was estimated at 18.1 million and 19.3 million, respectively \cite{bray2018global, sung2021global}. Consequently, it is of utmost importance to develop innovative methods for early-stage diagnosis of cancer. In recent years, computer-aided diagnosis (CAD) systems have become a popular tool to address this need by exploiting ML algorithms \cite{yassin_machine_2018}. However, the success of these algorithms is contingent upon the availability of large quantities of accurately labeled and annotated data. Acquiring such data can be both expensive and labor-intensive, leading to a scarcity of usable data. Additionally, these models often struggle with generalizability when applied to new data \cite{chao2021generalizing}.

To address the lack of accurate annotated data, Chen et al. \cite{chen2022generating} proposed a novel meta-learning-based contrastive learning method called Meta Ultrasound Contrastive Learning (Meta-USCL). The key idea of Meta-USCL was to properly generate and weigh positive sample pairs with a positive pair generation module and develop an automatic sample weighting to tackle the key challenge of obtaining semantically consistent sample pairs for contrastive learning. The Meta-USCL was evaluated for various medical problems, such as breast cancer classification, breast tumor segmentation, and pneumonia detection. Song et al. \cite{song2022diagnosis} proposed an interpretable structure-constrained graph neural network (ISGNN) to distinguish between pseudo-progression (PsP) and true tumor progression (TTP) in Glioblastoma multiforme (GBM) patients. The ISGNN used a meta-learning strategy for aggregating class-specific graph nodes to enhance classification performance on small-scale datasets while maintaining interpretability. Zhang et al. \cite{zhang2022development} proposed a multi-modal meta-learning model to detect small peritoneal metastases (PM) less than 0.5 cm in size using Computed Tomography (CT) images. To increase the classifier's robustness, the Maximum Mean Discrepancy (MMD) loss was utilized for updating the networks based on prior distributions. Chao et al. \cite{chao2021generalizing} developed a meta-learning framework for classifying whole-genome doubling (WGD) across 17 cancer types using digitized histopathology slide images. WGD is the most common phenotype among different types of cancers. However, identifying WGD using traditional methods is challenging due to tissue-specific variability and the limited availability of labeled data. The framework considers cancer types as independent jointly-learned tasks, leading to faster learning than traditional models.

\par To tackle the domain shift challenge, Li et al. \cite{li_utilizing_2022} proposed an unsupervised domain adaptation framework based on graph convolutional networks (GCN) and meta-learning for lethal pancreatic cancer segmentation. In the first stage, the proposed framework utilized encoders based on adversarial learning to transform source MRI images into a target-like visual appearance. This process ensured the segregation of domain-invariant features from domain-specific ones. The employed meta-learning strategy effectively coordinates between the original and transformed images, facilitating the exploration of associative features. In the subsequent step, a GCN network supervises the extracted abstract features related to segmentation. Achmamad et al. \cite{achmamad2022few} presented a meta-learning framework specifically tailored for brain tumor segmentation. The framework utilized information extracted from a limited number of annotated support images during episodic learning. This extracted information, in turn, guided the segmentation process of query images. In each episode, an encoder extracts feature maps for both support and query images. Then, they acquire guidance features through a process known as masked average pooling, which selectively considers only the target categories in the supported image. The relationship between the guidance features and the query feature maps is established through a convolutional operation. They demonstrated that their innovative methodology enhanced generalization capabilities in the area of few-shot semantic brain tumor segmentation, particularly when combined with a U-Net architecture designed around a decoder-based structure. A VGG-16 network, which had been pre-trained on ImageNet, was utilized as the encoder for both the support and query images to extract feature maps. Using a combination of DL and decision trees in a meta-learning framework, Lei et al. \cite{lei2022meta} introduced a meta-ordinal regression forest (MORF) model to address the generalization challenge and improve classification performance. This method includes a tree-wise weighting net and a grouped feature selection module. Utilizing a meta-learning approach as a recommender system to decide whether to use a shape or texture-based classifier, Byra et al. \cite{byra2022deep} improved the robustness of a neural network for the classification of breast cancer. Cao et al. \cite{cao2021auxiliary} introduced a self-adapting meta-learning model, which integrates with CL for skin cancer detection, even in the presence of noisy data. Initially, the model was trained on data representing common diseases and later adapted to classify rare skin conditions. This approach effectively reduced both training time and associated costs. Dubey et al. \cite{dubey2022neural} presented a meta-learning framework for brain tumor MRI image analysis. They introduced a neural augmentation technique that utilizes meta-learning to generate synthetic training samples with the aim of increasing the generalizability of the model.

\subsubsection{Dementia}
Dementia is a medical condition that manifests as a gradual deterioration in cognitive function, impairing an individual's ability to perform everyday tasks. This decline is primarily attributed to the damage of nerve cells \cite{alagiakrishnan2013evaluation}. Early symptoms of the onset/development of this condition may include losing track of time, forgetting recent events, experiencing difficulty in making decisions, and other related indicators. It's also common for dementia patients to experience anxiety, show diminished empathy, and exhibit frustration over their memory loss \cite{who-dementia}. According to global data, approximately 55.2 million people were living with dementia in 2019. Assuming the prevalence rate remains constant and factoring in population growth, projections suggest that by 2030 and 2050, the number of dementia patients could increase to around 78 million and 139 million, respectively. Moreover, the worldwide expenditure on this cognitive ailment was approximated to be US\$1.3 trillion and US\$2.8 trillion in 2019 and 2030, respectively \cite{world2021global}. The enormity of the disease burden highlights the urgent need for early detection and preventative measures to tackle dementia worldwide effectively. Recently, the application of meta-learning in the detection and diagnosis of cognitive disorders has gained significant attention. Alzheimer's disease (AD), in particular, the most prevalent form of dementia globally, is an area where such methods have been increasingly employed \cite{song2021auto, guan2021cost, han2023ml, zou2022meta}.

Song et al. \cite{song2021auto} presented an auto-metric graph neural network (AMGNN) model for AD diagnosis, wherein they incorporated a meta-learning strategy to facilitate inductive learning, enabling independent testing across various node classification tasks. The AMGNN model addressed the existing GNN algorithms' inability to balance flexibility with performance and the reduction of sample size caused by the incorporation of multimodal features. The AMGNN demonstrated the capability to manage structured and unstructured data as well as small sample sizes effectively.

\par Subjective cognitive decline (SCD) or subjective memory complaint (SMC) is a pathological condition that occurs prior to the prodromal phase of AD, which is referred to as Mild Cognitive Impairment (MCI). The detection of AD during the SCD phase faces several obstacles, including a low number of subjects, resulting in data scarcity and a lack of significant objective impairment in the brain. These challenges hinder the creation of a generalizable and robust model, making it difficult to develop reliable and adaptable methods for detecting AD during the SCD phase. To address the mentioned challenge, Guan et al. \cite{guan2021cost} introduced a cost-sensitive meta-learning framework for predicting the progress of SCD using MRI images. They introduced a key concept in their cost-sensitive learning approach; unlike conventional learning models, it assigned higher penalty values for misclassification, leading to enhanced sensitivity and reduced false negative rate.

\subsubsection{Cardiovascular Diseases}
Cardiovascular disease (CVD) refers to a broad spectrum of disorders affecting the heart and blood vessels. This term encompasses a variety of conditions, including coronary heart disease, cerebrovascular disease, peripheral arterial disease, and rheumatic heart disease. Recent studies and global disease statistics reveal that heart disease holds the grim distinction of being the leading cause of death worldwide, with an estimated 17.9 million fatalities in 2019 \cite{who_cvd_factsheet, vos_global_2020}. The impact of heart disease is not evenly spread across the globe, as it disproportionately affects low- and middle-income countries. Notably, around three-quarters of heart disease-related deaths occur within these regions \cite{who_cvd_factsheet}. This variation can be partly attributed to limited access to healthcare services and a higher incidence of risk factors. These risk factors include high blood pressure, elevated cholesterol levels, smoking habits, obesity, and a lack of physical activity \cite{world2022noncommunicable}.

\par ML algorithms possess a revolutionary capacity to transform CVD management by improving diagnostic precision, treatment approaches, and preventive tactics \cite{krittanawong2020machine,mathur2020artificial}. They can be meticulously trained to scrutinize medical images for the detection and diagnosis of CVDs, aiding healthcare professionals in formulating more precise diagnoses and designing highly effective treatment plans. These models can help pinpoint individuals at an elevated risk of developing CVD. They can significantly enhance prevention strategies and enable early intervention, potentially reducing the global burden of heart disease. However, domain shifts between the model's training and application phases and data scarcity are two common challenges in CVD diagnosis using ML models \cite{ahsan2022machine}. To address these issues, a meta-learning approach can be used to learn the model parameters in a transferable fashion. The model can be quickly adapted to the application phase using only a few training examples. \cite{upadhyay2021sharing}

\par To overcome the constraints posed by the limited availability of labeled data in medical imaging, Hansen et al. \cite{hansen2022anomaly} presented a novel approach to few-shot cardiac MRI image segmentation by leveraging self-supervision with supervoxels and drawing inspiration from anomaly detection techniques. The main idea behind the approach revolved around utilizing supervoxels to generate pseudo-labels for self-supervised learning. The algorithm first extracted superpixels from the input images and then deployed a DNN to learn a feature representation that could differentiate between normal and abnormal regions. This self-supervised learning process was guided by the supervoxel-based pseudo-labels, which were generated without manual annotation. The learned feature representation was then fine-tuned using a small number of labeled examples in a few-shot learning setting. Wibowo et al. \cite{wibowo2022cardiac} proposed a method for diagnosing heart disease, specifically cardiomyopathy patient groups, using only segmented output maps from cardiac cine MRI. They modified a fully convolutional EfficientNetB5-UNet encoder-decoder network for semantic segmentation of MRI image slices. They used a two-dimensional thickness algorithm to develop 2D representation images of the end-diastole and end-systole cardiac volumes. Then, they employed a few-shot learning model with an adaptive subspace classifier for classification.

\subsection{Acute Diseases}
Acute diseases usually arise suddenly and are characterized by severe and swift symptoms. This category of diseases can range from mild conditions to serious life-threatening events. The key defining characteristic of acute diseases, regardless of their severity, is the short duration of symptoms, typically from a few days to a few weeks. While acute diseases are typically short-lived, they may require immediate and serious medical attention \cite{linden2022prevalence}. In all cases, an accurate diagnosis is essential, usually involving a detailed examination and possibly further testing, to determine the exact nature of the acute condition and the best course of treatment \cite{vos_global_2020}.

\subsubsection{COVID-19}
In 2019, a novel pathogen named SARS-CoV-2 emerged, triggering a respiratory disease coined as COVID-19. This new disease swiftly propagated across the globe, prompting the WHO to categorize it as a pandemic \cite{who_covid19}. The manifestation of COVID-19 in individuals covers a spectrum of symptoms: fever, fatigue, loss of taste or smell, sore throat, myositis, and in severe instances, acute respiratory distress syndrome (ARDS). \cite{cdc_covid19_symptoms, paliwal2020neuromuscular} To date, this pathogen has been responsible for more than 6.8 million fatalities worldwide \cite{who_covid19, covid2020severe}. Efforts to diagnose COVID-19 have hinged largely on the analysis of patient medical data. A general diagnostic approach employs test kits. Despite their widespread use, these kits possess several limitations. Issues with false negatives and positives, accessibility in remote or underfunded regions, and the time required to return results present significant challenges to the efficacy and efficiency of these tests \cite{wang2021automatically, long2020diagnosis, ai2020correlation}. As such, the search for improved, rapid, and reliable diagnostic methods continues unabated.

\par ML algorithms have shown promising potential to help with diagnosing COVID-19 \cite{albahri2020role, panahi2021fcod, subramanian2022review}. In the early stages of the disease outbreak, ML models were mostly employed to detect the disease using medical images like chest X-ray (CXR) and CT scan images \cite{salehi2020coronavirus}. However, due to the limited availability of training data for DL algorithms, the model development process necessitated using pre-trained models. This posed two main challenges: If the features learned by the pre-trained models didn't align closely with the task, the performance and training time of the transfer learning were unsatisfactory. As such, meta-learning models started to develop to mitigate the impact of these challenges. 

Shorfuzzaman et al. \cite{shorfuzzaman2021metacovid} developed a meta-learning framework called MetaCovid for accurately diagnosing COVID-19 using CXR images with limited sample size. MetaCovid used a convolutional SNN with a pre-trained VGG-16 as a base model to enhance the feature embeddings obtained from input images. Chen et al. \cite{chen2021momentum} aimed to reduce the time required for annotating chest CT scans by proposing a self-supervised meta-learning framework. The meta-learning stage is leveraged to fine-tune the encoders used in representation learning. They employed contrastive learning to train an encoder capable of capturing rich feature representations from extensive lung datasets and adopted the network for classification. Miao et al. \cite{miao2021umlf} proposed an unsupervised meta-learning algorithm (UMLF-COVID) to address the challenges associated with domain shift and data scarcity in identifying COVID-19 patients from CXR images. Their model, a 4-layer neural network, was capable of identifying CXR images of COVID-19, Viral Pneumonia, Bacterial Pneumonia, and Normal individuals.

\subsection{Others}
While the two previous sections are devoted to topic-specific applications of meta-learning models with image inputs, other more general medical-related challenges perhaps have received less attention from researchers. One such domain is the investigation of healthcare-related texts. 

\subsubsection{Healthcare-related Texts}
Adopting natural language processing (NLP) models for healthcare-related applications is on the rise, driven by their recognized capability to search, process, analyze, and interpret a significant volume of data \cite{hudaa2019natural}. Advanced NLP techniques can extract insights from text-based information that was previously deemed inaccessible. Despite physicians spending a substantial amount of time documenting the reasons and details behind a patient's condition, this invaluable data is often overlooked to be considered as the input of the developed ML models. Moreover, the internet is a prolific data source that could be extracted and analyzed for healthcare-related insights. However, despite the progress in NLP-based ML and DL models, the efficacy of existing methods is contingent upon the quality, volume, and diversity of content types \cite{yin2020meta}. These characteristics are not usually available for all healthcare-related unstructured data.

\par Meta-learning based approaches are well-equipped to address this challenge. Ling et al. \cite{ling2022metagnn} introduced a graph meta-learning model for medical professional vocabulary representation. They also combined a GNN and LSTM model for learning unstructured medical professional discourse, which often includes entities with multiple meanings. They evaluated the performance of the proposed method on EHRs from 110 hospitals. Lu et al. \cite{lu2021novel} proposed a multi-modality fusion model for COVID-19 rumor detection on social media. Pirayesh et al. \cite{pirayesh2021mentalspot} presented the MentalSpot framework for early screening of depression on social media based on social contagion and meta-learning, where the implied knowledge at the individual level was transferred between friends. Wang et al. \cite{wang2022recognizing} developed a medical search query intent recognition model with few labeled data. The method used a query encoder that melds semantic data from a medical graph, syntactic insights, and knowledge from pre-trained language models.

\begin{table*}[]
\centering
\small
\caption{Summary of meta-learning applications in many/few-shot learning.}
\begin{tblr}{
colspec={p{1.55cm} p{1.45cm} p{0.65cm} p{1.6cm} p{1.45cm} p{0.75cm} p{0.85cm} p{1.25cm} p{0.8cm} p{4.1cm}},
  cell{2}{1} = {r=10}{},
  cell{2}{2} = {r=7}{},
  cell{9}{2} = {r=2}{},
  cell{12}{1} = {r=3}{},
  cell{12}{2} = {r=3}{},
  vline{2},vline{3},vline{4},vline{5},vline{6},vline{7},vline{8},vline{9},vline{10},
  hline{1-2,12,15-16} = {-}{},
  hline{3-8,10,13-14} = {3-10}{},
  hline{9,11} = {2-10}{},
}
\hline
\textbf{Domain}                             & \textbf{Application}             & \textbf{Ref.}           & \textbf{Strategy}           & \textbf{Methods}                         & \textbf{Num. of shots} & \textbf{Input type}                                  & \textbf{Data Acquisition}                                             & \textbf{Result$^{*}$} & \textbf{Highlights or Limits}                                                                                                                       \\
Chronic diseases and comorbidities & Cancer                  & \cite{chen2022generating}         & Optimization-based & Contrastive meta weight network & Many       & US                       & Butterfly, CLUST, Liver Fibrosis, COVID19-LUSMS           & 0.95 & Introduce a module for generating positive pairs, complemented by an automatic system for weighing samples.                                \\
                                   &                         & \cite{song2022diagnosis}         & Metric-based       & Graph generation constraints    & Few        & MRI & Wake Forest School of Medicine dataset                    & 0.83 & Develop an interpretable ISGNN for the automated distinction between PsP and TTP by focusing on meta-tasks related to various small graphs.       \\
                                   &                         & \cite{zhang2022development}        & Optimization-based & Customized loss functions       & Many       & CT                                          & 131 patients                                              & 0.87 & Enhance the model's generalization and make it accessible with minimal data requirements.                                                  \\
                                   &                         & \cite{li_utilizing_2022}           & Optimization-based & Meta-GCN                        & Many       & MRI                                         & Ruijin Hospital                                           & 0.61 & Tackle the coordination challenge between the source and transformed images.                                                               \\
                                   &                         & \cite{achmamad2022few}~           & Optimization-based & Meta-CNN                        & Few        & MRI                                         & BraTS2021~                                                & 0.65 & Segment tumor regions with limited labeled images.                                                                                         \\
                                   &                         & \cite{chao2021generalizing}                & Optimization-based & Meta-CNN~                       & Few~       & Histo pathology                              & TCGA                                                      & 0.70 & Mitigate batch-effect for real-time detection of cancer phenotypes.                                                                         \\
                                   &                         & \cite{cao2021auxiliary}          & Optimization-based & Meta-CNN                        & Many       & Skin biopsy images                          & ISIC2018                                                  & 0.79 & Introduce a self-modifying weighted model to increase the performance of skin cancer detection.                                            \\
                                   & Dementia                & \cite{song2021auto}         & Metric-based       & AMGNN                           & Many       & MRI + EHR                                   & TADPOLE                                                   & 0.94 & Address the lack of flexibility of existing GNNs in the face of multimodal data.                                                           \\
                                   &                         & \cite{guan2021cost}        & Optimization-based & Meta-CNN                        & Many       & MRI                                         & Two datasets                                              & 0.60 & Develop a cost-sensitive meta-learning approach to enhance the generalizability and sensitivity of an automated model for SCD detection.~ \\
                                   & CVD~ & \cite{wibowo2022cardiac}       & Optimization-based & Meta-CNN                        & Few        & MRI                                         & 2017 MICCAI~                                              & 0.92 & Handel the limited available data for classifying heart disease.                                                                           \\
Acute diseases                     & COVID-19                & \cite{shorfuzzaman2021metacovid} & Metric-based       & SNN                             & Many~      & CXR                                         &  Covid-chestxray-dataset, Kaggle-pneumonia                & 0.96 & Propose a faster model convergence and greater generalizability approach.                                                                  \\
                                   &                         & \cite{chen2021momentum}         & Optimization-based & Meta-CNN                        & Few        & CT                                          & COVID-19 CT, LIDC-IDR, ISMIR                              & 0.87 & Address the lack of annotated CT images during the initial stages of the pandemic for model development.                                     \\
                                   &                         & \cite{miao2021umlf}        & Optimization-based & Meta-CNN                        & Few        & CXR                                         & BIMCV‑ COVID19, Kaggle-pneumonia, Chest XRay\_AI & 0.90 & Construct an unsupervised meta-learning model to swiftly screen individuals suspected of COVID-19.                                         \\
Others                             & Healthcare related text & \cite{ling2022metagnn}          & Optimization-based & Meta-GNN                        & Few        & Text                                        & Reddit, GDKD                                              & 0.98 & Develop a model to learn medical professional vocabulary representation even with multiple meanings. \\                                   
\hline 
\end{tblr}
\begin{minipage}{\textwidth} 
\textsuperscript{*}{Accuracy is reported in classification studies, and DSC is listed in segmentation studies.}
\end{minipage}
\end{table*}

\subsection{Insights and implications}
The many/few-shot learning section reveals several analytical insights into the application of meta-learning in healthcare. By leveraging meta-learning in the few-shot scheme, models can effectively learn from just a few examples for a given task, enabling their deployment in data-scarce scenarios that are very common in medical domains. In many-shot learning, however, meta-learning provides a powerful way to transfer knowledge from data-rich sources to the target task with limited labeled examples to quickly adapt and boost performance. While meta-learning in these cases can be computationally expensive, some works explore formulations that are more efficient, including the cost-sensitive framework. Such considerations are important for scaling to larger medical datasets. Also, most research focuses on improving performance, with limited attention on making models more interpretable. Upon analyzing various studies, we observed a prevalent use of optimization-based algorithms across diverse input data types. Despite the potential of model- and metric-based methods, their application in healthcare has been limited so far. Moreover, the use of zero- and one-shot learning approaches is minimal despite their potential to serve as baseline models and provide insights into the generalizability of developed models. It should be noted that multimodal meta-learning approaches have yet to gain significant attention in healthcare.

\section{Development Highlights}

\subsection{Libraries}
Efficient implementation of meta-learning algorithms can benefit from the development of novel models built upon well-established machine-learning frameworks, such as TensorFlow \cite{abadi2016tensorflow} and PyTorch \cite{paszke2019pytorch}. Several libraries are designed explicitly for meta-learning and contribute to the field's rapid development (Tabel \ref{tab:Softwares}). At present, most of these libraries are actively being developed to optimize their scripts, add new meta-learning algorithms, and combine meta-learning with other ML and DL models. Also, more tools are in the pre-alpha phase of development. As Python is the predominant programming language for the development of ML models, most of these libraries are written in Python, except the Flux model zoo and mtlSuite, which are written in Julia and R, respectively. These libraries typically leverage the automated mechanisms provided by other libraries and packages to implement meta-learning algorithms.

\begin{table}[h!]
  \centering
\caption{Developed libraries for meta-learning}
  \label{tab:Softwares}
\begin{tabular}{  p{3cm} | p{2cm} | p{2cm} }
                            \hline \hline
    \textbf{Library}  & \textbf{Language} & \textbf{Framework}      \\  \hline
    
    BOML \cite{liu2021boml} & Python & TensorFlow    \\  \hline
    Flux model zoo \cite{innes2018flux} & Julia & Julia   \\  \hline
    higher \cite{grefenstette2019generalized} & Python & PyTorch    \\  \hline
    learn2learn \cite{arnold2020learn2learn} & Python & PyTorch   \\  \hline
    metagym & Python & PyTorch  \\  \hline
    MetaNN & Python & PyTorch    \\  \hline
    mtlSuite \cite{mantovani2019meta} & R & R    \\  \hline 
    pyMeta & Python & TensorFlow  \\  \hline
    Torchmeta \cite{deleu2019torchmeta} & Python & PyTorch    \\  \hline \hline

\end{tabular}
\end{table}

\subsection{Computation Cost}

Meta-learning strategies are primarily used where the number of data is insufficient to train a learning model in the healthcare domain; however, there is growing attention towards methods that seek to extend different meta-learning paradigms to more extensive datasets. While meta-learning models can potentially eliminate certain high computational cost steps, such as data augmentation, they often involve additional layers of optimization and more intricate learning processes. This can lead to increased resource demands, including memory and processing time, compared to traditional ML models \cite{huisman2021survey}. Also, when an objective-based meta-learning strategy is considered for a model, additional computational costs related to optimizing task-specific objectives may arise. To meet this challenge, numerous strategies have been developed to mitigate the computational costs of meta-learning models, with each solution offering different tradeoffs \cite{hospedales2021meta}. In general, the specific costs and tradeoffs associated with meta-learning models can vary depending on the strategy employed, and in most cases, the benefits of these models may outweigh their higher computational demands, particularly in domains where data is scarce or highly diverse.

\subsection{Benchmarks and Baselines}
Progress in the ML field is typically measured and spurred by benchmarks and is tangible compared to baseline models. ML benchmarks define a dataset(s) and task(s) to assess the generalization of a model over unseen instances. In contrast, baseline models are usually trained and tested on the same dataset. In essence, benchmarks set the stage for comparison, while baselines offer a specific reference point against which new models or techniques are measured.

Despite the development of several well-designed benchmarks for general-purpose meta-learning tasks \cite{bohdal2023meta,triantafillou2019meta}, there still remains a limited number of benchmarks specifically tailored for meta-learning in healthcare. FHIST is a highly diversified public benchmark compiled from various public datasets for few-shot histology data classification \cite{shakeri2022fhist}. The BSCD-FSL benchmark encompasses dermatology and radiology images for many/few-shot learning analysis \cite{guo2020broader}. A clinical meta-dataset derived from a data hub called The Cancer Genome Atlas Program (TCGA) provides 174 different tasks aimed at predicting multiple clinical variables using gene-expression data \cite{samiei2019tcga}. Additionally, datasets extracted from large clinical databases, such as metaMIMIC from the MIMIC-IV database, can serve as benchmarks for a broad spectrum of tasks \cite{woznica2023consolidated}. Moreover, researchers can utilize the various meta-learning models outlined in this paper as baseline studies. Comparing the performance of these models with other ML and DL models also provides a suitable choice for assessing meta-learning models in healthcare. Newly developed methods may also be evaluated against other state-of-the-art meta-learning models in a closely related field. More general potential baselines can be found here \cite{baum2022meta,sharaf2020meta, hu2023compressed}. However, it is crucial to consider those newly created algorithms designed for clinical domains might exhibit differing levels of interoperability, generalizability, bias, run time, computational cost, and memory usage compared to the baseline models. Therefore, using performance metrics as the sole criterion for comparison might not provide a complete picture in many studies.

\section{Discussion}

\subsection{Bias}
Factoring in bias while developing meta-learning models for healthcare applications is a significant aspect. Bias may not only arise at the model level of a meta-learning algorithm and during knowledge transfer, but it can also emerge in healthcare data. Bias in clinical decision support systems can lead to disparate medical treatment based on patient cohorts, which can directly result in patient harm \cite{spector2021respecting}. It can also cause extensive damage to the trust of individuals using these systems \cite{norori2021addressing}. Therefore, addressing bias is vital during the development process of such models in order to avoid significant repercussions.

\par At the model level, bias can occur when an algorithm produces systematically prejudiced outcomes because of flawed assumptions in the development process. While real-world implementation \cite{mccradden2020ethical}, open-source libraries \cite{beam2020challenges}, experimental methodology \cite{liu2016deepfood}, and end users \cite{tonekaboni2019clinicians} are general sources of biases in ML and DL algorithms that should be considered, meta-learning algorithms may face additional sources of bias. Since meta-learning models can demonstrate superior performance in recognizing patterns with a minority population of data, they are a primary choice when the available data is scarce. In few-shot and zero-shot learning approaches, the developed model tends to be biased towards the observed classes during the training phase \cite{chao2016empirical}. In heterogeneous problems, such as clinical risk prediction, meta-learning models may not exhibit outstanding performance across all metrics, and bias can arise in the engineered features, and the model evaluation phases \cite{dhiman2022risk}. These issues become more critical when developing models that are provided with imbalanced or unrepresentative data \cite{lee2019learning, zhao2019cost}. As a result, a meta-learning model may represent a non-generalizable way to learn from this experience.

\par Despite providing an effective transferable knowledge scheme for different healthcare applications, a meta-transfer learning approach may not guarantee sufficient transferring of critical knowledge from a source domain. Therefore, a non-negligible bias toward the source model exists even after model adaptation to the target domain \cite{salman2022does}. The transferred bias could arise in realistic settings (e.g., pre-training on ImageNet or other standard models) if the target dataset is considered to be explicitly unbiased \cite{salman2022does}. In addition, meta-learning approaches may encourage developers to randomly generate a simulated target task and adapt their models to numerous tasks \cite{vinyals2016matching}. However, since transfer learning models often require capturing domain shifts, the simulated target used cannot be blindly sampled. While task distribution is considered one of the decisive factors in the success of meta-learning models, it could be a reason for biases \cite{zhang2019metapred}. As such, the diversity of transferring knowledge between the investigated source and target domains should be appropriate. Alternatively, supplementary algorithms and methods should be applied to mitigate the effects of any potential biases.

\subsection{Model Validation}
A common practice in developing meta-learning models, as with other learning schemes, is to perform sample splitting, partitioning the dataset into train and test (validation) sets. As previously discussed, meta-learning models are a primary choice for situations where the available data for developing a learning model is limited. Although sample splitting is generally believed to be necessary for evaluation, it may have potential drawbacks regarding data efficiency since neither the training nor the testing phase can exploit sufficient data. Bai et al. \cite{bai2021important} demonstrated that train-test splitting is unnecessary when a task is generated from noiseless linear models. Moreover, the train-test split is not optimal in the realizable setting, but it is essential in the general agnostic setting. Alternatively, models could be trained using all available data and subsequently evaluated on another task distribution. This strategy is valid as long as it allows for reliable model selection and a consistent performance comparison across different methods \cite{setlur2021two}. Additionally, the discussed key algorithms, particularly those from various categories, are designed to enhance different facets of model development, such as generalizability, latency, and accuracy, which makes them incomparable on the basis of a single metric alone \cite{vanschoren2018meta, hospedales2021meta, tian2022meta}. As a result, studies comparing these approaches should validate models against a broader spectrum of performance criteria. 

\subsection{Generalizability}
Generalizability refers to the ability of a model to perform well on unseen data across a range of tasks, conditions, and populations \cite{yang2022machine}. In healthcare, this is crucial as data is often heterogeneous, and a model's performance may vary between patient cohorts or clinical settings. One of the main challenges to the generalizability of meta-learning models in healthcare is the diversity of tasks and contexts for various diseases and conditions. The curse of dimensionality is another challenge that affects the generalizability of meta-learning models. As the number of dimensions or features in the data increases, the training data required to cover the feature space grows exponentially, while the available training data is oftentimes limited in healthcare. In such situations, meta-learning models may face challenges in balancing learning task-specific information and acquiring transferable knowledge across tasks.

\par To improve the generalizability of meta-learning models in healthcare, several strategies can be employed. Incorporating domain knowledge into the learning process can help guide the model toward more meaningful representations and adaptations. This could involve integrating expert knowledge, such as clinical guidelines or diagnostic criteria, into the model's architecture or loss function. Doing so can help the model focus on relevant features and adapt more effectively to new tasks. The generalizability of meta-learning models can be enhanced by employing meta-regularization and adversarial training techniques. Meta-regularization involves adding penalties to the model's loss function to encourage the learning of more general features, while adversarial training involves exposing the model to a range of challenging tasks or data perturbations during training. Another strategy to learn invariant features is to use multi-task learning frameworks. In multi-task learning, the model is trained on multiple related tasks simultaneously, encouraging the learning of shared representations across the tasks. By jointly optimizing the model for various tasks, the model can identify and leverage the common features that contribute to better generalization across different problem domains.

\subsection{Interpretability}
Despite the limited preliminary success, most ML approaches cannot extract interpretable knowledge and procedure from this data deluge. The majority of DL-based approaches, including meta-learning ones, function as a black box, receiving a set of data as inputs and directly outputting a result that is difficult to interpret. In spite of remarkable success in tackling the challenges associated with clinical support systems development, the lack of interpretability might lead to rendering the outcomes unable to reveal the correlation between different features of the output, imposing specific desirable properties, such as safety constraints or worst-case guarantees, for further debugging and improvement \cite{yu2021reinforcement}. While the outcome of meta-learners might not generate a set of rules and indicators (humanly interpretable) that formulate a scoring system or diagnostic criteria, the interpretability limitation could hinder the successful adoption of meta-learning methods for clinical applications, as physicians are unlikely to accept new approaches without rigorous safety validation and correctness \cite{lipton2017doctor,shickel2017deep}.

\par Various approaches can be considered to increase the interpretability of meta-learning in different stages of model development. A meta-learner could be more interpretable by integrating concrete instances and high-level meta-information \cite{dong2020meta}, constructing a meta-hierarchical tree to capture the structured information \cite{yao2020online}, considering prior knowledge or constraints to provide physically consistent predictions \cite{karniadakis2021physics}, utilizing regularization schemes to encourage orthogonality of nominal and learned features \cite{banerjee2020adaptive}, using a GP algorithm to encode the model in evolutionary trees \cite{meqdad2022meta}, establishing a meta-knowledge graph to discover the most relevant structure and tailor the learned knowledge to the learner \cite{yao2020automated}, visually explaining the outcome \cite{wang2021mtunet}, replacing a black box classifier with decision tree \cite{zhang2022metadt}, indicating the direct relationship between final classification and region similarities \cite{xue2020region}, or investigating the features' influences \cite{shao2022find}. Nonetheless, meta-learners' interpretability in the healthcare domain is not thoroughly addressed and calls for further investigations.

\subsection{Other Challenges and Open Issues}

The initial efforts and progress in employing meta-learning within the healthcare sector over the previous years were discussed. Significant attention has been directed toward a wide range of application areas within healthcare. Despite noteworthy accomplishments, most studies to date have largely depended on basic meta-learning methods to confront healthcare challenges. This section focuses on general challenges that current research has yet to adequately tackle.

\subsubsection{Hyperparameter Sensitivity} Meta-learning models often have many hyperparameters, and their performance can be highly sensitive to the settings of these hyperparameters \cite{vanschoren2018meta}. This can make these models challenging at the meta-level to tune on healthcare data (characterized by a high dimensionality of features) and apply in practice.

\subsubsection{Task Distribution} Meta-learning commonly assumes a unimodal distribution of tasks, suggesting that a single learning strategy would suffice for all tasks. However, this assumption is often not met in practical scenarios, which results in suboptimal performance \cite{tseng2020cross,sun2019meta,balaji2018metareg,li2019feature}.  The complex reality is that different tasks within the distribution may demand diverse learning strategies, a challenge not easily addressed with contemporary methods. Despite the success of meta-learning within specific task families in the healthcare domain \cite{tan2022metacare++,singh2023meta,samiei2019tcga}, better methods for handling shifts specifically designed for different types of healthcare data are needed.

\subsubsection{Evaluation Protocols} Evaluation protocols for meta-learning applications in healthcare are limited, overly simplified, or unrealistic \cite{singh2023meta,van2022meta}. More rigorous and realistic evaluation protocols are needed to effectively assess meta-learning models' performance. Also, meta-learning lacks universally accepted benchmarks for different applications and data types.

\subsubsection{Scalability} Meta-learning models often are complex to scale because they involve nested optimization problems (optimizing the learning rule, then optimizing within each task) \cite{simon2020modulating,acar2021memory,kamath2019eml}. As the number or complexity of tasks increases, the computational resources required for training the meta-learning model can scale dramatically. Additionally, there is a risk of overfitting when the number of data or tasks increases, not just for the training scheme but also for the meta-training procedure \cite{chen2022understanding}. 

\subsubsection{Task Defining and Encoding} The definition and encoding of a task play a crucial role in meta-learning. Poor task definitions can make learning difficult while limiting the ability to transfer learned knowledge to new tasks \cite{zou2022meta}. Following this, determining which parts of the model contributed to the final performance is a non-trivial problem and can affect the interpretability and transparency of the model.

\subsubsection{Integration with Other Learning Paradigms} Meta-learning is not the only advanced learning paradigm. Other methods like active learning, unsupervised learning, semi-supervised learning, and self-supervised learning also aim to improve the learning process. Integrating meta-learning with those paradigms is a promising but open research direction, especially in healthcare.

\section{Future Perspectives}
After discussing previous studies and outlining the major challenges and open problems of meta-learning in healthcare, this section focuses on the future perspectives of meta-learning development for healthcare applications. We explore three primary trajectories: the prospective utilization of meta-learning in various healthcare contexts, the success of meta-learning algorithms in distinct fields that have received limited attention in healthcare, and the ways to the practical use of a meta-learning model in a real-world setting.

\subsection{Application}
Though meta-learning models have been widely applied in numerous healthcare domains for various tasks (from prediction and diagnosis to treatment and management), their use has not been widely adopted for a substantial range of medical conditions and diverse types of input data. While other ML models have been developed and tailored to serve as decision support systems for various medical conditions (such as sepsis, ARDS, human immunodeficiency virus (HIV), anemia, depression, and anesthesia) and different medical intervention tasks (like ventilation, dosing, and heparin administration), the development of meta-learning models has been relatively limited. The development and deployment of meta-learning algorithms to address these issues might present an interesting direction for enhancing the capabilities and performance of intelligent models.

\par  While the inherent appropriateness of applying meta-learning techniques to omics-related data and its potential applications is apparent, the diversity and number of models developed thus far are still limited. The complexities in the interplay of omics data, compounded by the scarcity of available samples, present a significant challenge \cite{li2022machine,reel2021using}. Nonetheless, meta-learning models promise to be a potent tool for overcoming the existing constraints in this field.

\par The applications of meta-learning methods call for further investigations in certain types of medical data. Areas such as medical text, healthcare-related questionnaires, and question-answering systems could greatly benefit from meta-learning. Additionally, meta-learning is seldom applied to different image data types like histopathology images, even though these are typically available in limited quantities and are apt for domain shift applications. Moreover, the majority of meta-learning models developed to work with medical images primarily focus on classification tasks. However, attention to other tasks (such as segmentation and recognition) remains scarce in the healthcare domain. Auditory input data represents another significant area that is unexplored using meta-learning. This includes utilizing sounds from body organs to detect abnormalities and employing speech recognition techniques for diagnosing mental disorders.

\par Epidemic prediction may significantly benefit from meta-learning \cite{wang2022deep,moon2023model,panagopoulos2021transfer}. By training on diverse past epidemics, meta-learning algorithms can discern common patterns, enhancing predictions for future outbreaks. These models can adapt quickly to new epidemics using few-shot learning to make accurate predictions early on, even when data is limited.

\subsection{Algorithm}
As thoroughly discussed  above, most of the applied meta-learning methods in the healthcare domain have been limited to certain algorithms. This presents a significant opportunity to assess the effectiveness of the other meta-learning-based algorithms. Also, it has rarely been found that the performance of several meta-learning models was evaluated and compared for a single healthcare challenge. While numerous ML architectures are available for various medical tasks, the number of meta-learning models specifically proposed for these tasks is notably limited. Furthermore, bio-inspired meta-learning approaches have not yet garnered significant attention. By combining these two concepts, bio-inspired meta-learning seeks to create learning algorithms that not only learn how to learn but do so in a way that emulates the processes found in biological systems.

\par Physics-informed ML models are proposed as a promising solution to the scarcity of data and could increase the performance of ML models by incorporating mathematical physics models \cite{karniadakis2021physics,meng2022physics}. With the progress in the modeling of healthcare conditions, physics-informed meta-learning may be an interesting topic to explore \cite{sameni2020mathematical,cobelli2019introduction}. Moreover, the effectiveness of numerous novel meta-learning methods that represent significant results in different fields, such as online meta-learning \cite{finn2019online}, meta continual learning \cite{vuorio2018meta}, meta-gan \cite{bao2022storm}, meta federated learning \cite{aramoon2021meta}, self-supervised meta-learning \cite{kedia2021keep}, unsupervised meta-learning \cite{khodadadeh2019unsupervised}, zero-shot learning \cite{wang2019survey}, and multimodal meta-learning \cite{vuorio2019multimodal}, remains largely unexplored within the healthcare sector. 

\par Applying meta-learning to reinforcement learning (RL) tasks can lead to more efficient policies for patient treatment, prognosis prediction, and resource allocation. While a significant number of RL models have been developed to address healthcare challenges \cite{yu2021reinforcement}, the exploration of meta-RL models in this sector remains relatively untouched. Given that the available training data is typically limited in domains where an RL model may be the optimal choice - particularly in situations where the generation of synthetic or simulated data fails to replicate the characteristics of real-world data - meta-RL has the potential to deliver promising results. Moreover, meta-learning methods driven by human knowledge could be of significant value, whereas conventional learning algorithms may falter because of complex and incomplete data or an inexplicable learning process. Therefore, integrating humans (medical practitioners) into the learning procedure and aligning expert knowledge with learning data could substantially enhance the knowledge discovery process. While there has been some preceding work from other sectors \cite{chen2023fast,wan2020human}, the concept of human-in-the-loop interactive meta-learning that can learn from human feedback is still not well established in the healthcare field.

\section{Conclusion}
Meta-learning presents an adept approach to making optimal decisions in a range of healthcare scenarios by leveraging learning to learn. This paper aims to deliver extensive, state-of-the-art of meta-learning applications to a wide range of challenges in the healthcare setting. We have prepared a structural summary of the theoretical underpinnings and pivotal techniques in meta-learning research from a traditional machine learning viewpoint. We surveyed the wide-ranging applications of meta-learning methods in tackling challenges within a multitude of healthcare domains, including multi/single-task learning and many/few-shot learning. A detailed discussion has been provided on the development and challenges within the current research landscape of meta-learning in healthcare. These discussions have been undertaken from the perspectives of the fundamental components that constitute a meta-learning process to key issues in meta-learning research in healthcare. While significant strides have been made in addressing the complex issues within the broader meta-learning community, applying these solutions directly to healthcare contexts can present unique difficulties due to the inherent complexity of medical data processing. In essence, the unique features exhibited by the medical decision-making process necessitate the creation of advanced meta-learning models genuinely suited for real-world healthcare problems. Furthermore, we have reviewed the overlooked domains, such as bias, interpretability, and generalizability.

\par The applications of meta-learning in healthcare is a multi-disciplinary field involving computer science and medicine. Thus, this interdisciplinary research necessitates concerted efforts from machine learning researchers and clinicians involved directly in patient care. Meta-learning has achieved notable successes, especially in the challenges such as the insufficient amount of data and domain shifts that other learning paradigms struggle with by leveraging prior knowledge; however, it still hasn't received as much attention as machine learning and deep learning methods in healthcare. Based on the rapid and significant progress of meta-learning techniques and practical requirements from healthcare practitioners, an escalation in enthusiasm now exists for applying meta-learning in healthcare. As the first comprehensive survey of meta-learning in healthcare, we hope to provide the research community with a systematic understanding of meta-learning foundations, available methods and techniques, potential applications, existing challenges, and future perspectives. We believe this will inspire more researchers from various disciplines to apply their expertise collaboratively, paving the way for more applicable solutions for optimal decision-making in healthcare. Our aim is to encourage researchers from different fields to apply their individual expertise and collaborate to produce more practical solutions for healthcare challenges.
\newline

\par \textbf{Statements and Declarations}
\par \textbf{Competing Interests}
\par The authors declare that they have no conflict of interest.
\newline
\par \textbf{Funding Information}
\par Not Applicable
\newline
\par \textbf{Author Contribution}
\par AR, RM, FH, and RK conceived the initial idea for the survey, outlining its scope, foundational concept, and structure of the paper. AR, RM, and SJ did the literature review process, with RK providing oversight. AR, RM, SJ, and FH write the main manuscript. AR, FJ, and RK provided critical manuscript review and revision. RM and SJ had equal contributions. All authors reviewed the manuscript. All authors read and approved the final manuscript.
\newline
\par \textbf{Data Availability Statement}
\par Not Applicable
\newline
\par \textbf{Research Involving Human and /or Animals}
\par Not Applicable
\newline
\par \textbf{Informed Consent}
\par Not Applicable
\newline
\par \textbf{Acknowledgments}
\par Not Applicable

\bibliographystyle{IEEEtran}
\bibliography{Mendeley}
\end{document}